\documentclass[preprint,12pt]{elsarticle}

\usepackage{amssymb}
\usepackage{tabularray}
\usepackage{url}
\usepackage{multirow}
\usepackage{longtable}
\usepackage{booktabs,multicol,multirow,tabularx,array} 
\usepackage{makecell}
\usepackage{float}
\usepackage{comment}
\usepackage{amsmath,amsfonts}
\usepackage{afterpage}
\usepackage[T1]{fontenc}
\usepackage{hyperref}
\usepackage{natbib}
\usepackage{graphicx}
\usepackage[subrefformat=parens,labelformat=parens]{subfig}
\usepackage{cleveref}
\usepackage{siunitx}
\makeatletter %to remove preprint submitted to blah blah
\def\ps@pprintTitle{%
  \let\@oddhead\@empty
  \let\@evenhead\@empty
  \let\@oddfoot\@empty
  \let\@evenfoot\@oddfoot
}
\makeatother

\begin{document}

\begin{frontmatter}

%% Title, authors and addresses

%% use the tnoteref command within \title for footnotes;
%% use the tnotetext command for theassociated footnote;
%% use the fnref command within \author or \address for footnotes;
%% use the fntext command for theassociated footnote;
%% use the corref command within \author for corresponding author footnotes;
%% use the cortext command for theassociated footnote;
%% use the ead command for the email address,
%% and the form \ead[url] for the home page:
%% \title{Title\tnoteref{label1}}
%% \tnotetext[label1]{}
%% \author{Name\corref{cor1}\fnref{label2}}
%% \ead{email address}
%% \ead[url]{home page}
%% \fntext[label2]{}
%% \cortext[cor1]{}
%% \affiliation{organization={},
%%             addressline={},
%%             city={},
%%             postcode={},
%%             state={},
%%             country={}}
%% \fntext[label3]{}

%\title{Extracting Diagnosis Pathways from Electronic Health Records using Deep Reinforcement Learning}
\title{Deep Reinforcement Learning for Personalized Diagnostic Decision Pathways Using Electronic Health Records: A Comparative Study on Anemia and Systemic Lupus Erythematosus }

% use optional labels to link authors explicitly to addresses:
% \author[label1,label2]{}
% \affiliation[label1]{organization={},
%             addressline={},
%             city={},
%             postcode={},
%             state={},
%             country={}}
%
% \affiliation[label2]{organization={},
%             addressline={},
%             city={},
%             postcode={},
%             state={},
%             country={}}

\author[inst1,inst2]{Lillian Muyama\corref{cor1}}

\affiliation[inst1]{organization={Inria Paris},%Dept and Org
            % addressline={2 Rue Simone IFF}, 
            city={Paris},
            postcode={75012}, 
            % state={State One},
            country={France}}

\author[inst1,inst2,inst3]{Antoine Neuraz}
\author[inst1,inst2]{Adrien Coulet}
\cortext[cor1]{Corresponding author: \url{lillian.muyama@inria.fr}}

\affiliation[inst2]{organization={Centre de Recherche des Cordeliers, Inserm, Université Paris Cité, Sorbonne Université},
            % addressline={Address Two}, 
            city={Paris},
            postcode={75006}, 
            % state={State Two},
            country={France}}
            
\affiliation[inst3]{organization={Hôpital Necker, Assistance Publique - Hôpitaux de Paris},
            % addressline={Address Two}, 
            city={Paris},
            postcode={75015}, 
            % state={State Two},
            country={France}}

% \author[1,2]{Lillian Muyama\corref{cor1}%
% \fnref{fn1}}
% \ead{lillian.muyama@inria.fr}
% \author[1,2,3]{Antoine Neuraz\fnref{fn2}}
% % \ead{antoine.neuraz@aphp.fr}
% \author[1,2]{Adrien Coulet\fnref{fn1,fn3}}
% % \ead{adrien.coulet@inria.fr}
% \cortext[cor1]{Corresponding author}
% % \fntext[fn1]{This is the first author footnote.}
% % \fntext[fn2]{Another author footnote, this is a very long
% % footnote and it should be a really long footnote. But this
% % footnote is not yet sufficiently long enough to make two
% % lines of footnote text.}
% % \fntext[fn3]{Yet another author footnote.}
% \affiliation[1]{organization={Inria Paris},%Dept and Org
%             % addressline={2 Rue Simone IFF}, 
%             city={Paris},
%             postcode={75012}, 
%             % state={State One},
%             country={France}}
% \affiliation[2]{organization={Centre de Recherche des Cordeliers, Inserm, Université Paris Cité, Sorbonne Université},
%             % addressline={Address Two}, 
%             city={Paris},
%             postcode={75006}, 
%             % state={State Two},
%             country={France}}
% \affiliation[3]{organization={Hôpital Necker, Assistance Publique - Hôpitaux de Paris},
%             % addressline={Address Two}, 
%             city={Paris},
%             postcode={75015}, 
%             % state={State Two},
%             country={France}}
\begin{abstract}
%% Text of abstract
\textbf{Background}: Clinical diagnosis is typically reached by following a series of steps recommended by guidelines authored by colleges of experts. Accordingly, guidelines play a crucial role in rationalizing clinical decisions but suffer from limitations as they are built to cover the majority of the population and fail at covering patients with uncommon conditions. Moreover, their updates are long and expensive, making them unsuitable for emerging diseases and practices. 
\newline
\textbf{Methods}: Inspired by guidelines, we formulate the task of diagnosis as a sequential decision-making problem and study the use of Deep Reinforcement Learning (DRL) algorithms to learn the optimal sequence of actions to perform in order to obtain a correct diagnosis from Electronic Health Records (EHRs). We apply DRL on synthetic, but realistic EHRs and develop two clinical use cases: Anemia diagnosis, where the decision pathways follow the schema of a decision tree; and Systemic Lupus Erythematosus (SLE) diagnosis, which follows a weighted criteria score. We particularly evaluate the robustness of our approaches to noisy and missing data since these frequently occur in EHRs.
\newline
\textbf{Results}: In both use cases, and in the presence of imperfect data, our best DRL algorithms  exhibit competitive performance when compared to the traditional classifiers, with the added advantage that they enable the progressive generation of a pathway to the suggested diagnosis which can both guide and explain the decision-making process.
\newline
\textbf{Conclusion}: DRL offers the opportunity to learn personalized decision pathways to diagnosis. We illustrate with our two use cases their advantages: they generate step-by-step pathways that are self-explanatory; and their correctness is competitive when compared to state-of-the-art approaches. 

\end{abstract}

%%Graphical abstract
% \begin{graphicalabstract}
% \includegraphics[scale=0.3]{images/revised_drawio_jbi_graphical_abstract.png}
% \end{graphicalabstract}

%%Research highlights
% \begin{highlights}
% \item The development of clinical practice guidelines is mostly manual, driven by experts.
% \item Clinical practice guidelines and clinical pathways can be represented in various formats.% depending on their purpose and the resources available.
% \item This systematic review provides an in-depth look at approaches to develop clinical practice guidelines and pathways.
% \item No study uses data-driven or hybrid (expert- and data-driven) methods to guide the clinical practice guideline development process.
% \item We claim that data-driven methods could guide the experts and make their developments faster.
% \end{highlights}

\begin{keyword}
%% keywords here, in the form: keyword \sep keyword
Diagnosis \sep Decision support \sep Deep Reinforcement Learning \sep Decision pathways \sep Anemia \sep Lupus
%% PACS codes here, in the form: \PACS code \sep code
% \PACS 0000 \sep 1111
%% MSC codes here, in the form: \MSC code \sep code
%% or \MSC[2008] code \sep code (2000 is the default)
% \MSC 0000 \sep 1111
\end{keyword}

\end{frontmatter}

%% \linenumbers

%% main text

\section{Introduction}\label{sec:intro}
For complex diagnosis decisions, clinicians customarily refer to diagnosis guidelines, which define the successive steps one should take to reach a diagnosis. 
These documents are typically established by a college of experts based on the best available evidence \cite{field1990clinical}, and rationalize and normalize clinical decisions.
The succession of steps they recommend includes information collection, observations, laboratory test orders, and other exams. 
However, clinical guidelines suffer several drawbacks. First, they are designed to cover the majority of the population and hence may fail to guide to the right diagnosis in the case of uncommon patients such as elders, or those with multiple diseases. 
Second, their establishment is long and expensive, and updates are usually done after several years \cite{steinberg2011clinical}. 
This makes them unsuitable to fast emerging practices such as those associated with a recently developed laboratory test or an emerging disease, such as the recent COVID-19 pandemic. 
Moreover, as clinical guidelines are expensive and time-consuming to produce, their development does not scale to the full spectrum of diseases. 
More versatile and scalable methods are therefore required to provide insights when clinical guidelines are incomplete, or not available.

We think that machine learning approaches trained on clinical data may complement diagnosis guidelines. In particular, we aim at developing approaches able to guide the decision process in a step-by-step manner, as described in \cite{adler2021}. We believe that such an approach may reduce the number of irrelevant tests therefore optimizing healthcare costs, but primarily may propose more personalized and accurate diagnoses, especially in the case of patients with uncommon conditions. 

The collection of patient-level data in Electronic Health Records (EHRs) offers great opportunities to gain knowledge about clinical practice \cite{jensen2012}. 
EHRs encompass structured, semi-structured, and unstructured data about patients' health such as medications, laboratory test orders and results, diagnoses, as well as demographic features. 
Previous works relied on EHRs to train machine learning (ML) methods to automatically suggest diagnoses for patients such as in \cite{lipton2015learning, miotto2016, choi2016doctor}. However, in these studies, supervised ML methods are employed to predict a unique endpoint, \textit{i.e.}, the diagnosis represented as a class label.
We believe that for data-driven approaches to find adoption in clinical practice, it is important for a diagnosis not to be limited to an endpoint, but to be represented as a pathway that follows steps of medical reasoning and decision-making. 

In this work, we propose to leverage EHR data and investigate how it can be used to train deep reinforcement learning (DRL) models to build explainable pathways supporting the diagnosis of complex conditions. We develop our approach on two primary use cases: the diagnosis of anemia and Systemic Lupus Erythematosus (SLE), commonly referred to as Lupus. 
Anemia is a clinical condition defined as a lower-than-normal amount of healthy red blood cells in the body, which we chose for three reasons: its diagnosis is made mainly based on a series of laboratory tests that are available in most EHRs; it is a common diagnosis implying that the associated amount of data may be sufficient to train RL models; and the differential diagnosis of anemia is frequently complex to establish, making its guidance useful.
SLE is a chronic autoimmune disease where the immune system attacks the body's health tissues. 
We selected SLE because its diagnosis is complex to establish and because the sequence of steps recommended to reach the diagnosis follows a schema that differs from anemia. In anemia, recommendations mainly follow the schema of a decision tree \cite{bmj_anemia}, whereas recommendations for SLE diagnosis are based on a weighted criteria score, covering an alternate decision schema.

We choose to use RL because it builds a model that can be used sequentially, passing through various steps named \emph{actions} and \emph{states} to reach a final objective state, which is similar to our objective. We propose here an adaptation of the framework of RL to construct individualized pathways of observations in a step-by-step manner, in order to suggest a diagnostic decision. 
For example, in the anemia use case, a pathway is a sequence of laboratory test requests (actions), whose results are then obtained (states) before either requesting for another test or terminating on a diagnosis. 
We make the assumption that the constructed pathways can complement clinical guidelines to aid practitioners in decision-making during the diagnosis process. 

Our main contributions are:
\textit{(i)}
    an adaptation of the RL framework to progressively construct  %individualized 
    optimal sequences of actions to perform in order to reach a diagnosis and
\textit{(ii)} 
    an empirical analysis on synthetic EHRs that illustrates the validity of our approach, identifies the most suitable algorithms, and evaluates their robustness with regard to varying levels of missing and noisy data. 

The rest of the paper is structured as follows. Section \ref{sec:related_work} presents related works. Section \ref{sec:methods} details the methods we deploy in this paper, including subsection \ref{sec:cohort} that explains the construction of our synthetic datasets. Section \ref{sec:results} presents the results of our empirical analysis, evaluating these methods in various scenarios. Section \ref{sec:discussion} discusses our findings and their limitations. Finally, Section \ref{sec:conclusion} concludes the study and provides directions for future work.

\section{Related Work}\label{sec:related_work}
Previously, studies have used various process mining, machine learning and statistical methods to extract clinical pathways from medical data. 
Zhang \textit{et al.} proposed Markov Chains for the identification of clinical pathways using patient visits data~\cite{zhang2015paving}. Perer \textit{et al.} extracted common patterns from data in order to build pathways~\cite{perer2015mining}, while Baker \textit{et al.} used a Markov model to extract meaningful clinical events from patient data~\cite{baker2017process}. 
Others adopted a statistical approach to build treatment pathways by using a Latent Dirichlet Allocation-based method~\cite{huang2013latent, huang2014discovery, huang2018probabilistic}. %to create treatment pathways. 
Machine learning methods such as Long Short-Term Memory and RL have also been used to construct pathways for optimal treatment in several studies such as in \cite{lin2021personalized} and \cite{li2022electronic}, respectively. 
However, their focus is on building pathways for \emph{treatment} decision, whereas we aim at building pathways for \emph{diagnosis}.

Likewise, numerous studies have leveraged Machine Learning (ML) methods for diagnosis. Given the longitudinal nature of EHRs, a prevalent choice has been Recurrent Neural Networks (RNNs) \cite{lipton2015learning, choi2016doctor}. Additionally, RNNs have also been used in the prediction of future patient outcomes such as in \cite{koshimizu2020prediction}. 
In \cite{obaido2022interpretable, zoabi2021machine, kavya2021machine}, novel efforts are made to not only employ ML approaches to provide clinical diagnoses but also to provide explanatory elements to enable the interpretation and contextualization of the results of the model.
However, our aim is not only to diagnose the patient with the correct condition but also to find the optimal sequence of features to observe to reach this diagnosis for each patient, which can be seen, in turn, as elements of explanation for the decision. In other terms, we aim to construct personalized diagnosis pathways that delineate the steps leading to the diagnosis, thereby explaining the diagnostic process.

Previous works used RL methods for costly feature acquisition in classification tasks \cite{li2021active, janisch2019classification}. However, these works did not primarily focus on diagnosis. Yu \textit{et al.} aimed to optimize the financial cost of several clinical processes, including the prediction of Acute Kidney Injury, using RL~\cite{yu2023deep}. While it was shown that their method reduced the cost of diagnosis, the actions taken at each step were not shown, therefore neglecting the relevant and explainable dimensions of pathways. Other studies used RL to diagnose patients by inquiring about the presence of disease-related symptoms from users~\cite{tang2016inquire, wei2018task, kao2018context}. While our approach similarly formulates the diagnosis process as a sequential decision-making problem and applies deep reinforcement learning (DRL) techniques, these studies use a symptom self-checking approach, whereas we aim to use EHR data. We believe that using EHRs is more suitable as they encompass data collected routinely in clinical practice, including objective and normalized measurements such as laboratory results therefore composing a great resource for training machine learning models and obtaining data-driven insights to aid clinical decision-making.

\section{Methods}\label{sec:methods}
\subsection{Decision Problem}
We consider the clinical diagnosis process as a sequential decision-making problem and formulate it as a Markov Decision Process (MDP) \cite{Littman2001markov}, following the RL \cite{sutton2018reinforcement} paradigm. 
Accordingly, we define an \emph{agent} interacting with an \emph{environment} in order to maximize a cumulative reward signal. 
At each time step $t$, the agent receives an observation $o$ of the \emph{state} of the environment, takes an \emph{action}, and obtains a \emph{reward}.  
The goal is to learn a policy, \textit{i.e.}, a function that maps states to actions, that maximizes the reward signal. 
In this study, our agents are taking observations from synthetic EHRs of a Clinical Data Warehouse (CDW), 
in order to reach a diagnosis, which is a final action. 

Let $\mathcal{D}$ denote a dataset of size $n \times (m+1)$ composed of $n$ patients, $m$ features, plus one diagnosis. $F$ is the set of the $m$ feature names; and $C$ the set of possible diagnosis values. An instance $D^i$ in $\mathcal{D}$ is a pair $(X^i, Y^i)$ with $X^i = \{x^i_1,\dots, x^i_j, \dots , x^i_m\}$ where $x^i_j$ is the value of the feature $j$ for the patient $i$, $m$ is the total number of features and $Y^i \in C$ is the anemia diagnosis of patient $i$.
Accordingly, our MDP is defined by the quadruple  $(S, A, T, R)$  as follows:
\begin{itemize}
    \item $S$ is the set of states. %In our problem, a state $s_t$, comprises the values of the features that have already been queried by the agent at time $t$ and is represented as a vector of fixed size $n$. 
    At each time step, the agent receives an observation $o$ of the state $s_t$, which is a vector of fixed size $m$ comprising the values of the features that have already been queried by the agent at time $t$; 
    features that have not been queried yet are associated with the value -1. Equation \ref{eqn:state_eqn} 
     defines the $j^{th}$ element of $o$, where $F'$ denotes the set of features that have already been queried by the agent. 
   \begin{equation}
    \label{eqn:state_eqn}
      o_j = \begin{cases}
      \;x_j, & \text{if $f_j \in F' $}\\
      -1, & \text{otherwise.}
    \end{cases}
    \end{equation}
    
    \item $A$ is the set of possible actions, which is the union of the set of \emph{feature value acquisition actions}, $A_f$ (or feature actions for short),  and the set of 
    \emph{diagnostic actions}, $A_d$. 
     At each time step, the agent takes an action $a_t \in A$. 
     Actions from $A_f$ are taking values from the set of features $F$. Accordingly a specific value $f_j \in F$ will trigger the action of querying for the value of this feature to the CDW.
     Actions from $A_d$ are taking values from the set of possible diagnoses $C$. 
     At a time step $t$, if $a_t \in A_d$, the episode is terminated. Also, for \textit{anemia}, if an episode reaches the number of maximum specified steps without reaching a diagnosis, it is terminated.
    
    \item $T$ denotes the transition function that gives the probability of moving from a state $s_t$ to a state $s_{t+1}$ given an agent action $a_t \in A$.

    \item $R$ is the reward function. $r_{t+1}$, which can be written as $R(s_t, a_t)$, is the immediate reward when an agent in a state $s_t$ takes an action $a_t$. In the case of a diagnostic action $a_t \in A_d$, if the diagnosis is correct, the reward is +1, %else,
    otherwise the reward is -1. 
    \begin{equation}
    \label{eqn:diag_actions}
       \text{if}\ a_t \in A_d, R(s_t, a_t) = \begin{cases}
      \;\;\,1, & \text{if $a_t = Y^i $}\\
      -1, & \text{otherwise}
    \end{cases}
    \end{equation}
    
Considering the \textbf{anemia} use case, for a feature action $a_t \in A_f$, if the feature has already been queried, the agent receives a penalty of -1 and the episode is terminated. 
Otherwise, the agent receives a reward of 0. Accordingly, the reward function for the feature actions is formalized as follows:

     \begin{equation}
     \label{eqn:anem_feat_actions}
      \text{if}\ a_t \in A_f, R(s_t, a_t) = \begin{cases}
      -1, & \text{if $a_t \in F' $}\\
      \;\;\,0, & \text{otherwise}.
    \end{cases}
    \end{equation}

Considering the \textbf{lupus} use case, for a feature action $a_t \in A_f$, if the feature has already been queried, the agent receives a penalty of -1 and the episode is terminated. Otherwise, the agent incurs a penalty determined by the penalty weight associated with the action (See Subsection \ref{sec:cohort}).
%Table \ref{tab:feature_scores}. 
The reward function is: 

    \begin{equation}
     \label{eqn:lupus_feat_actions}
      \text{if}\ a_t \in A_f, R(s_t, a_t) = \begin{cases}
      -1, & \text{if $a_t \in F' $}\\
      \frac{-1}{\lambda c}, & \text{otherwise},
    \end{cases}
    \end{equation}

where $c$ is the penalty weight of the feature action and $\lambda \in \mathbb{R}$ is a scaling factor.
\end{itemize}

\subsection{DQN and its extensions}
\textbf{Q-learning} \cite{watkins1992q} is an RL algorithm that outputs the best action to take in a given state based on the expected future reward of taking that action in that particular state. The expected future reward is named the Q-value of that state-action pair,
noted $Q(s,a)$. At each time step, the agent selects an action following a policy $\pi$, and the goal is to find the optimal policy $\pi^*$ that maximizes the reward function. 
During model training, the Q-values are updated using the Bellman Equation as described in equation \ref{eqn:q_learning} .

In our use case, since the state space is large and continuous, we propose to use a \textbf{Deep Q-Network (DQN)} \cite{mnih2015human} which uses a neural network to approximate the Q-value function. 
In order to improve DQN stability and performance, several extensions of the DQN algorithm have been developed. 
Particularly \textbf{Double DQN (DDQN)}, \textbf{Dueling DQN} and \textbf{Prioritized Experience Replay (PER)}, which we test in the subsequent sections of this paper and briefly describe in \ref{apd:Q}. 
The stability gained from these techniques enhances the performance by making the process faster and reducing overestimation bias. %These techniques enhance the performance of the DQN algorithm by improving its stability, making the learning process faster and more efficient, and reducing overestimation bias, among other benefits. 

\subsection{State-of-the-Art Classifiers}
As part of the study, we compared the performance of the DRL models with four traditional supervised learning algorithms that are commonly used for classification tasks, namely Decision Tree (DT), Random Forest (RF), Support Vector Machine (SVM), and a Feed-Forward Neural Network (FFNN). 
DT has a particular status in our study for two reasons. First, in the anemia use case, the labels of the synthetic dataset were assigned according to a decision tree. For this reason, the DT approach is expected to perform very well on that data. Second, DT is the only considered classifier that is self-explanatory, as the path in the DT that is taken to classify an instance constitutes a pathway to the diagnosis.

\subsection{Evaluation Approach and Implementation}
80\% of the dataset was used to train the model while the remaining 20\% was used as the test set. Additionally, 10\% of the training dataset (\textit{i.e.}, 8\% of the dataset) was used for validation. Since one aim of this study is to learn diagnosis pathways for anemia and lupus, and to evaluate how methods are robust to noise and missing data, the validation and test sets used in all the experiments are constant (\textit{i.e.}, without noise or missingness). 

Ten runs were conducted for the first experiment comparing DQN approaches and five runs for the subsequent experiments on the models' robustness. Each run was performed with the same datasets but with different seeds, which influenced the models' interactions with the environment, as well as initialisation. %This deliberate variation in seeds ensured a comprehensive exploration of the models' behavior under diverse initial conditions, enhancing the robustness and reliability of our findings. 

The metrics used to measure the performance of the models are:
\begin{itemize}
    \item \textbf{Accuracy}: The ratio of episodes that terminated with a correct diagnosis. %This was our primary metric. It gauged the model's overall correctness in its diagnostic predictions.
    \item \textbf{Mean episode length}: The average number of actions performed by the agent per episode. %It essentially measures the average length of the pathways generated by the model. 
    This metric does not apply to the SOTA classifiers that consider all the features. %due to their comprehensive consideration of all features in the classification process.
    \item \textbf{F1 score}: The harmonic mean of the precision and recall scores. We report one-vs-rest and macro-averaging F1 score.
    \item \textbf{ROC-AUC (Receiver Operating Characteristic - Area Under Curve) score}: It measures the model ability to distinguish between the different classes. We also used a one-vs-rest approach and macro-averaging.
\end{itemize}

Additional metrics were used in the \textbf{lupus} use case, as described below:
\begin{itemize}
     \item \textbf{Average pathway score (APS)}: The average score of a pathway is computed by summing the penalty weight of each feature leading up to the diagnosis. Formally,
     %Each individual pathway's score is computed using Equation \ref{eqn:pathway_score} where $p$ represents the length of the pathway, and $n$ is the total number of features in our dataset. This score is derived from the penalty weights outlined in Table \ref{tab:feature_scores}, by summing up the individual score of each feature leading up to the diagnosis in the pathway. 

     \begin{equation}
            s(p) =  1 - \frac{\sum_{i=1}^{|p|-1} w_i}{\sum_{i=1}^{m} w_i}
        \label{eqn:pathway_score}
    \end{equation}
    it is the score of a pathway $p$ where $w_i$ represents the penalty weight of the $i^{th}$ feature of $p$, $|p|$ is the length of $p$, and $m$ is the total number of features. 
    % The penalty weights of features used in the lupus use case are outlined in Section \ref{sec:cohort}.
    % appendix Table \ref{tab:feature_scores}.
    
    Equation \ref{eqn:avg_pathway_score} computes the average pathway score for the entire model:

     \begin{equation}
       \bar{s} = \frac{1}{n}\sum_{i=1}^{n}s(p_i)
        \label{eqn:avg_pathway_score}
        \end{equation}
        
    where $p_i$ is the $i^{th}$ pathway generated by the model and $n$ is the total number of pathways generated. %In our results, it is represented as a percentage.     
        
    \item \textbf{wPAHM (weighted Pathway score and Accuracy Harmonic Mean)}: we define wPAHM as the weighted harmonic mean of the accuracy and average pathway score of a model. This metric provides a unique value that reflects both the accuracy and average pathway score, according to their relative importance in a study context. Formally, 
    
    \begin{equation}
        \text{wPAHM} = \frac{w_a + w_{\bar{s}}}{\frac{w_a}{a} + \frac{w_{\bar{s}}}{\bar{s}}}
        \label{eqn: wpahm_score}
    \end{equation}

    where $w_a$ and $w_{\bar{s}}$ denote the weights for the accuracy and average pathway score, respectively.  In our case, we consider the accuracy of the diagnosis of higher priority than the score of our pathways. We therefore assigned weights of 0.9 and 0.1 to the accuracy ($w_a$) and average pathway score ($w_{\bar{s}}$), respectively. 
\end{itemize}

We used the OpenAI Gym Python library \cite{brockman2016openai} to implement our environment. For model training, we created an instance of the environment using the training data. At the start of each episode, the environment was reset using an instance drawn randomly from the training data. 
To build our agents, we used the stable-baselines package \cite{stable-baselines}, and the values of the hyperparameters are as shown in Table \ref{tab:hyperparams} in \ref{apd:general}. They were chosen based on prior knowledge, existing literature, and experimentation. 
As for the number of timesteps, for the \emph{anemia} use case, this was determined in the first run (using a grid-search strategy) using the validation data and remained constant for the other nine runs, while for the \emph{lupus} use case, all the models were trained using a constant number of timesteps, and a checkpoint for the model was saved at constant intervals. Using the validation data, the model checkpoint with the best performance was selected.
The hyperparameters for the SOTA models were chosen using a grid search strategy. The source code is available at \url{https://github.com/lilly-muyama/Deep_RL_diagnosis_pathways}.

\subsection{Datasets}\label{sec:cohort}

To experiment with our method we built two synthetic datasets, one for each use case (\textit{i.e.,} anemia and lupus). The details of the selection of the features, their values, the labeling of the diagnostic classes, and the inclusion of various levels of noise and missingness are detailed in \ref{apd:data}.

\begin{figure*}[htbp]
\includegraphics[width=\textwidth]{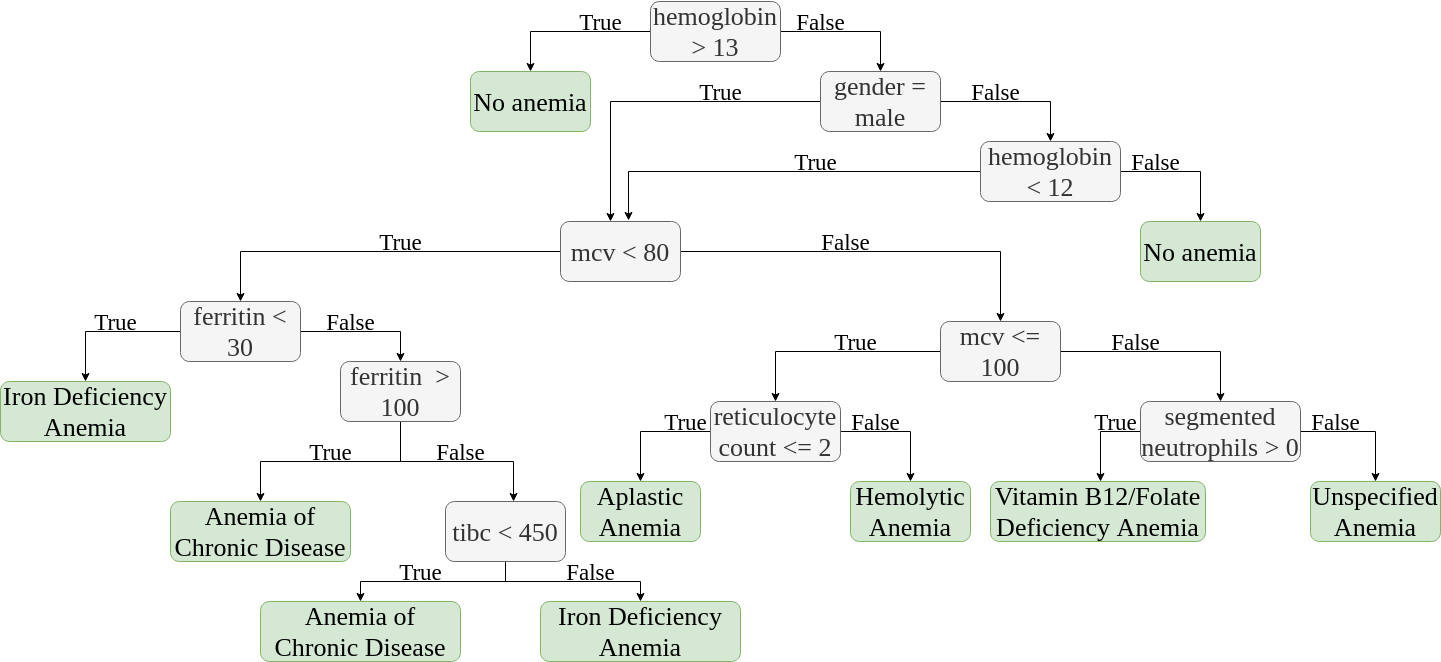}
\caption{The decision tree used to label our anemia dataset, adapted from \cite{bmj_anemia} and \cite{short2013iron}.}
\label{fig:tree}
\end{figure*}

The main difference in the construction of the two datasets is the rationale used to label synthetic patients with their diagnoses. For anemia, we rely on the DT depicted in Figure \ref{fig:tree}, whereas, for lupus, we rely on the weighted criteria score system defined in \cite{aringer20192019}. 
Here, the entry criterion for lupus diagnosis is the presence of Antinuclear antibodies (ANA) at a titer of $\geq$ 1:80 on HEp-2 cells.  The rest of the features are divided into several categories and only the highest criteria weight within a category is counted towards the patient's total weighted criteria score. If the patient's criteria weight is $\geq$ 10, the patient is diagnosed with Lupus. For example, if a patient is experiencing both delirium (criteria weight=2) and seizure (criteria weight=5), only 5 will be added to the patient's total score since both features belong to the neuropsychiatric category.

More Information regarding the datasets' characteristics and relevant details about the use cases such as the lupus features' penalty weights are presented in \ref{apd:anemia} and \ref{apd:lupus} respectively.

\section{Results}\label{sec:results}
\subsection{Performance Comparison of DQN and its extensions}
In the first round of experiments, we trained DQN, Double DQN, Dueling DQN, and Dueling Double DQN on our datasets; and also enabled prioritized experience replay (PER) for each of these models. In addition, we trained a Proximal Policy Optimization (PPO) algorithm using a similar number of timesteps for comparison purposes. 
%The associated results are compiled in Table \ref{tab:dqn_results}. 
Model names associated with the suffix -PER correspond to the mentioned DQN extension together with PER.
In order to gauge the stability of each model, we conducted ten different runs of the same experiment using different seeds, which results are reported in Table \ref{tab:dqn_results}. 

In both use cases, Dueling DQN-PER provided the best accuracy, while Dueling DDQN-PER exhibited a similar performance with slightly lower accuracy. In addition, Dueling DDQN-PER had the lowest standard deviation in the anemia use case and the second lowest in the lupus use case. Remarkably, it also demonstrated a significantly shorter mean episode length than the Dueling DQN-PER in the lupus use case. For these reasons, we used only these two DQN models for the rest of the experiments.
It should be noted that there are a few slight differences in the methodologies used in the two use-cases as seen in section \ref{sec:methods}.

\begin{table*}[!ht]
\centering
    \begin{tabular}{|p{3.7cm}|p{2.5cm}|p{2.5cm}|p{2.45cm}|p{2.45cm}|}
    \hline
    
    \multirow{2}{*}{\textbf{Model}} &
    \multicolumn{2}{c|}{\textbf{Accuracy}} &
    \multicolumn{2}{c|}{\textbf{Mean episode length}}\\
    \cline{2-5}
    & \textbf{Anemia} & \textbf{Lupus} & \textbf{Anemia} & \textbf{Lupus}\\\hline
         PPO & 25.58 $\pm$ 16.54 & 88.06 ± 7.19 & 1.63 ± 0.85 &  3.81 ± 1.63 \\ %& 12 Million \
         DQN & 89.31 $\pm$ 14.68  & 88.57 ± 4.89 &  4.27 ± 0.45 & 14.82 ± 3.17\\
         DDQN & 92.79 $\pm$ 5.37 & 90.97 ± 1.56 &  4.66 ± 0.31 & 11.42 ± 2.68 \\ % & 10 Million  \\
         Dueling DQN & 92.46 $\pm$ 5.31 & 97.08 ± 0.72 & 4.70 ± 0.24 & 22.57 ± 1.05 \\ % & 10 Million \\
         Dueling DDQN & 77.15 $\pm$ 31.99 & 94.20 ± 2.05 & 4.44 ± 1.02 & 16.28 ± 3.75 \\ % & 15 Million \\
         DQN-PER & 93.71 $\pm$ 5.51 & 94.17 ± 5.18 & 4.67 ± 0.29 & 20.46 ± 2.52  \\ % & \textbf{15 Million}\\
         DDQN-PER & 95.57 $\pm$ 2.01  & 94.26 ± 2.79 & 4.92 ± 0.36 & 13.27 ± 2.64\\ % &  15 Million \\
         {\small\textbf{Dueling DQN-PER}} & \textbf{96.64$\pm$ 1.46} & \textbf{98.81± 0.32} & 4.59 ± 0.20 & 22.09 ± 0.85 \\
          % &  15 Million \\15 Million \\
         {\small Dueling DDQN-PER} & 96.33$ \pm$ 1.32 & 98.39 ± 0.41 & 4.82 ± 0.45 & 14.40 ± 2.49 \\\hline
    \end{tabular}
    \caption{Performance of the RL models. Metrics are average and standard deviation over 10 runs.}%
    \label{tab:dqn_results}
  
\end{table*}

\subsection{Performance Comparison with SOTA classifiers}
The results of these experiments are shown in Table \ref{tab:sota_results}.
For the \emph{anemia} use case, since the diagnosis labels of the synthetic dataset were assigned following the decision tree in Figure \ref{fig:tree}, the tree-based agent, which is based on the same tree achieved a perfect score. This tree was not applicable to the \emph{lupus} use case. 
Also, we built an agent that acts randomly at each timestep for comparison purposes. Because the SOTA classifiers use a constant set of features to make a diagnosis, they do not have a mean episode length. The tree-based algorithms and the FFNN performed better than the DQN models, while SVM had a similar performance in the \emph{lupus} use case but had a lower performance in the \emph{anemia} use case. 

\begin{table*}[!ht]
\centering
  
    \begin{tabular}{|p{3.5cm}|p{2.5cm}|p{2.6cm}|p{2.45cm}|p{2.45cm}|}
    \hline
    \multirow{2}{*}{\textbf{Model}} &
    \multicolumn{2}{c|}{\textbf{Accuracy}} &
    \multicolumn{2}{c|}{\textbf{Mean episode length}}\\
    \cline{2-5}
    & \textbf{Anemia} & \textbf{Lupus} & \textbf{Anemia} & \textbf{Lupus}\\\hline
         Random Agent & 12.34 $\pm$ 0.33 & 17.69 ± 0.38 & 1.53 ± 0.01 & 4.75 ± 0.02\\
         Tree-based Agent & \textbf{100\ \ $\pm$ 0.00} & N/A & 3.98 ± 0.00 & N/A\\
         Decision Tree & 99.96 $\pm$ 0.00 & 99.42 ± 0.01 & N/A & N/A \\
         Random Forest & 99.90 $\pm$ 0.01 & 99.38 ± 0.02 & N/A  & N/A  \\
         FFNN & 97.97 $\pm$ 0.27 & \textbf{99.92± 0.03} & N/A  & N/A \\
         SVM & 94.89 $\pm$ 0.00 & 99.54 ± 0.00 & N/A & N/A \\
         {\small Dueling DQN-PER} & 96.64 $\pm$ 1.46 & 98.81 ± 0.32 & 4.59 ± 0.20  & 22.09 ± 0.85 \\
         {\small Dueling DDQN-PER} & 96.33 $\pm$ 1.32 & 98.39 ± 0.41 &  4.82 ± 0.45 & 14.40 ± 2.49 \\\hline
    \end{tabular}
    \caption{Performance of the DQN and the state-of-the-art classifiers. The tree-based agent has a perfect score because it acts according to the decision tree used to build the \emph{anemia} dataset.}
    \label{tab:sota_results}
\end{table*}

\subsection{Performance comparison using different metrics}
For the \emph{lupus} use case, since accuracy is not the only desirable metric, we also selected the models with the best performance according to an alternate metric, \textit{i.e.} the wPAHM in Equation \ref{eqn: wpahm_score}. The comparison results between the models with the best accuracy vs wPAHM scores are shown in Table \ref{tab:metrics_results}.

\begin{table}[!ht]
\centering
\begin{tabular}{|p{2.1cm}|p{1.6cm}|p{1.3cm}|p{2cm}|p{1.1cm}|p{1.5cm}|p{1.2cm}|p{1.6cm}|}
\hline
\textbf{Model} & \textbf{Goal metric} & \textbf{Acc.} & \textbf{Avg. ep. length} & \textbf{F1} & \textbf{ROC-AUC} & \textbf{APS} & \textbf{wPAHM}\\ 
\hline
\multirow{2}{*}{\makecell{Dueling\\ DQN-PER}} & Accuracy & \textbf{98.81} & 22.09 & \textbf{98.82} & \textbf{98.82} & 10.24 & 50.98 \\
\cline{2-8}
& wPAHM & 95.83 & 17.30 & 95.82 & 95.83 & 30.06 & 77.93 \\
\hline
\multirow{2}{*}{\makecell{Dueling\\ DDQN-PER}} & Accuracy & 98.39 & 14.40 & 98.39 & 98.39 & 43.15 & 86.18 \\
\cline{2-8}
& wPAHM & 96.41 & 9.56 & 96.42 & 96.42 & \textbf{63.64} & \textbf{91.59} \\
\hline
\end{tabular}%}
\caption{The mean performance of the best-performing DQN models for the \emph{lupus} dataset, based on a chosen goal metric (accuracy or wPAHM); with $\lambda$=9 in the reward function.}

\label{tab:metrics_results}
\end{table} 

% \begin{table}[!ht]
% \centering
% % \resizebox{\textwidth}{!}{
% % \begin{tabular}{|l|l|l|l|l|l|l|l|}
% \begin{tabular}{|p{1.4cm}|p{1.7cm}|p{1.7cm}|p{1.3cm}|p{1.5cm}|p{1.5cm}|p{1.5cm}|p{1.6cm}|}
% \hline
% \textbf{Model} & \textbf{Goal metric} & \textbf{Accuracy} & \textbf{Mean episode length} & \textbf{F1} & \textbf{ROC-AUC} & \textbf{Average pathway score} & \textbf{wPAHM}\\ 
% \hline
% Dueling DQN-PER & Accuracy & \textbf{98.81} & 22.09 & \textbf{98.82} & \textbf{98.82} & 10.24 & 50.98 \\
% \hline
% Dueling DQN-PER & wPAHM & 95.83 & 17.30 & 95.82 & 95.83 & 30.06 & 77.93 \\
% \hline
% Dueling DDQN-PER & Accuracy & 98.39 & 14.40 & 98.39 & 98.39 & 43.15 & 86.18 \\
% \hline
% Dueling DDQN-PER & wPAHM & 96.41 & 9.56 & 96.42 & 96.42 & \textbf{63.64} & \textbf{91.59} \\
% \hline
% \end{tabular}%}
% \caption{The mean performance of the best-performing DQN models for the \emph{lupus} dataset, based on a chosen goal metric (accuracy or wPAHM); with $\lambda$=9 in the reward function.}
% \label{tab:metrics_results}
% \end{table} 

\subsection{Results with different values of $\lambda$}
\begin{figure}[!ht] 
    \centering
    \subfloat[][]{\includegraphics[width=0.45\textwidth]{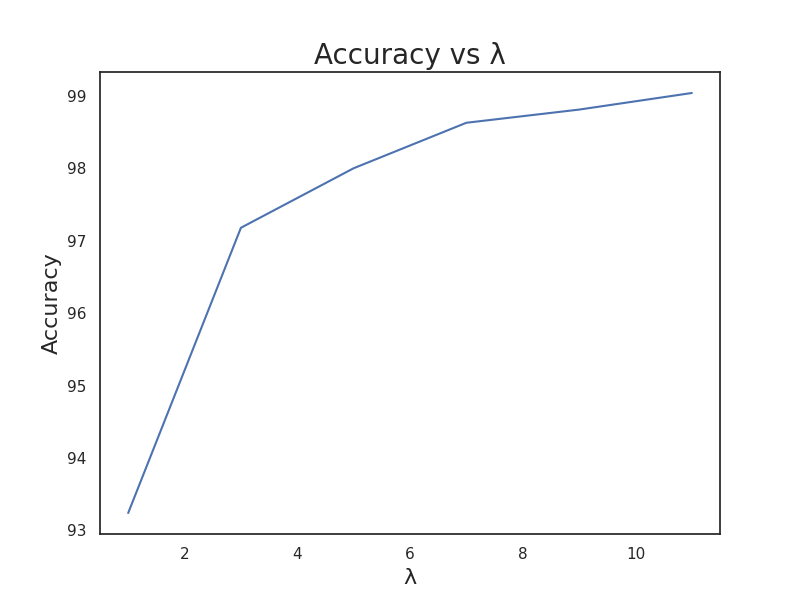}\label{varying_beta_acc}} 
    \subfloat[][]{\includegraphics[width=0.45\textwidth]{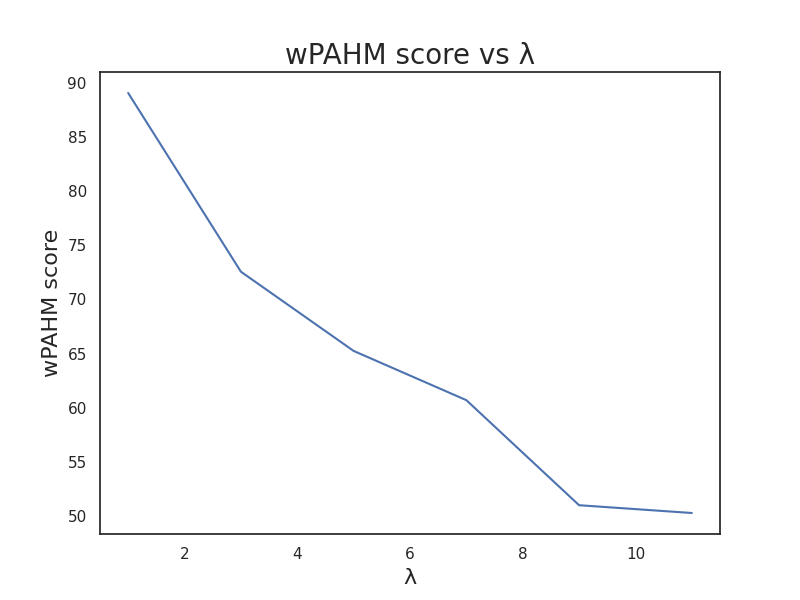}\label{varying_beta_wpahm}} 
    \caption{Graphs showing the effect of the $\lambda$ value in the reward function on \protect\subref{varying_beta_acc} the accuracy and \protect\subref{varying_beta_wpahm} the wPAHM scores for the dueling DQN-PER model.}
    \label{fig:varying_beta}
\end{figure}

In the reward function for the feature value acquisition actions in the \emph{lupus} use case (shown in Equation \ref{eqn:lupus_feat_actions}), the $\lambda$ acts as a scaling factor. We performed experiments with different values for $\lambda$, and the mean accuracy and wPAHM scores are illustrated in Figures ~\subref*{varying_beta_acc} and ~\subref*{varying_beta_wpahm} respectively. 
As shown, with the increase in the $\lambda$ value, the accuracy score rises while the wPAHM score declines. 

\begin{figure}
    \centering
    \includegraphics[scale=0.5]{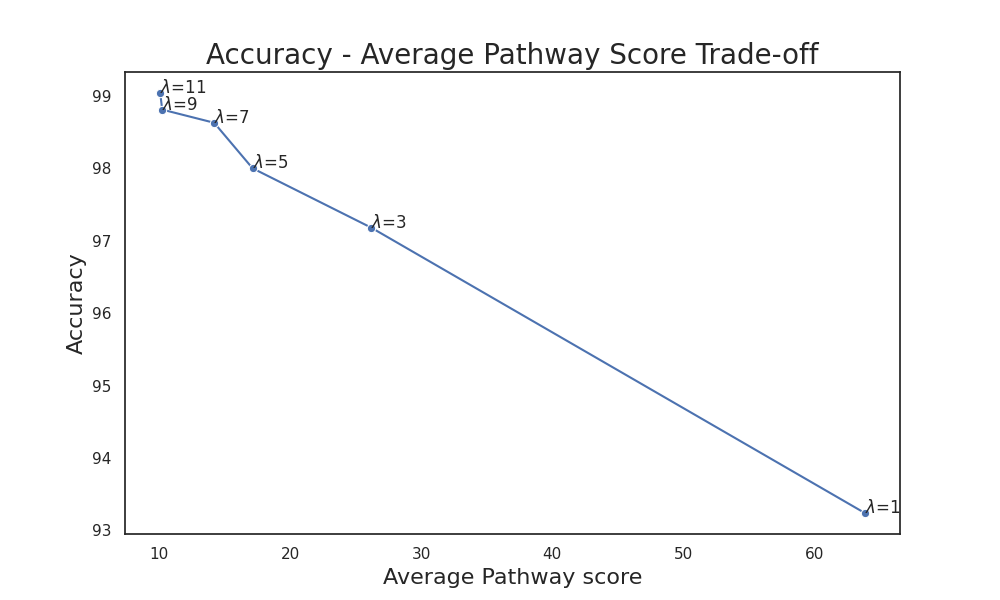}
    \caption{The accuracy-average pathway score tradeoff based on different values of $\lambda$ in the reward function. The results for the dueling DQN-PER model are used here.}
    \label{fig:acc_path_tradeoff}
\end{figure}

Additionally, Figure \ref{fig:acc_path_tradeoff} illustrates the trade-off between the accuracy and the average pathway score for different values of $\lambda$.

\subsection{Varying levels of missing data and noise}
\begin{figure*}[t!]
    \subfloat[]{%
        \includegraphics[width=.49\linewidth]{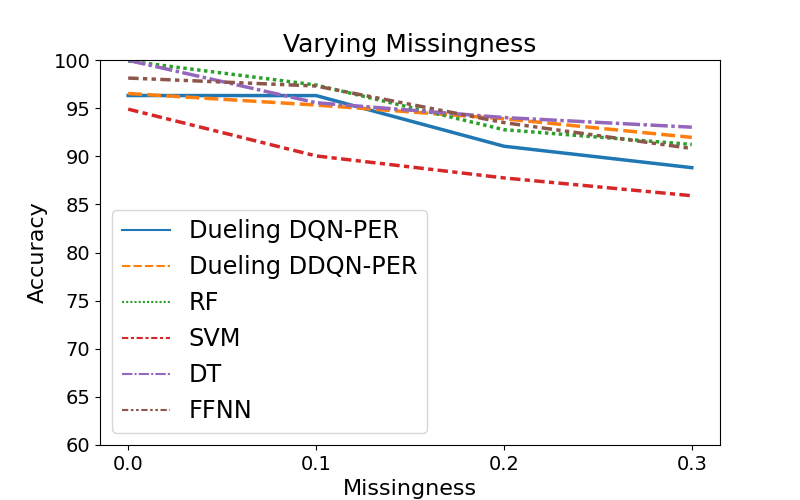}%
        \label{anem_missingness}%
    }\hfill
    \subfloat[]{%
        \includegraphics[width=.49\linewidth]{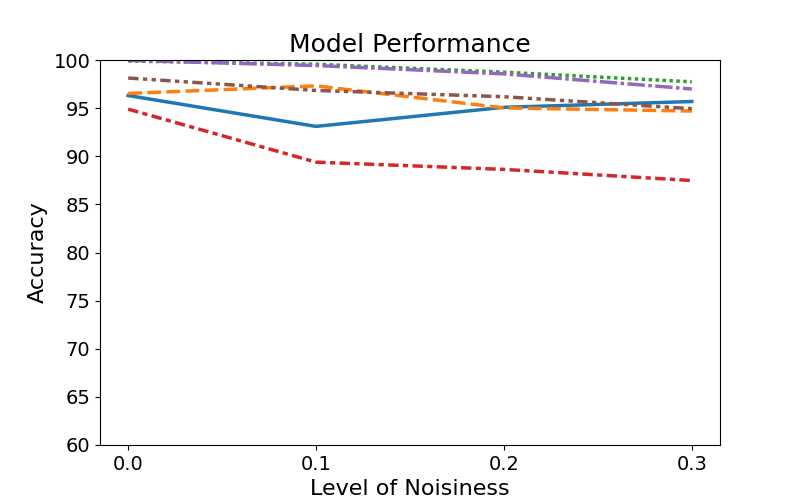}%
        \label{anem_noisiness}%
    }\\
    \subfloat[]{%
        \includegraphics[width=.49\linewidth]{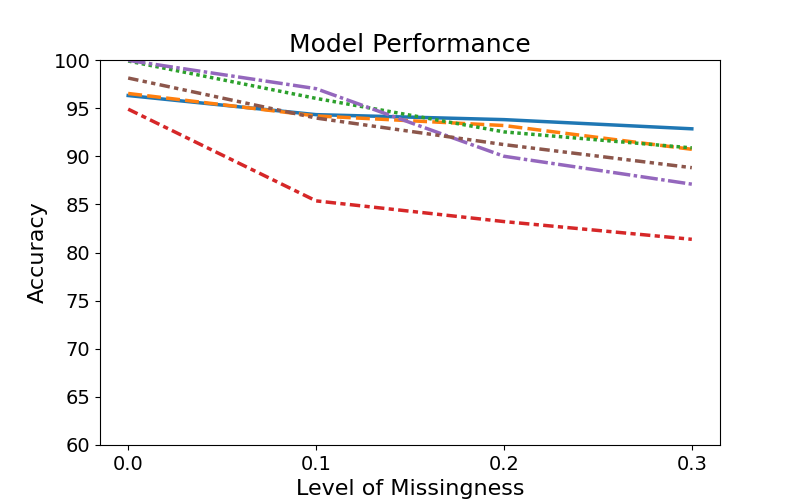}%
        \label{anem_missingness_noisiness}%
    }\hfill
    \subfloat[]{%
        \includegraphics[width=.49\linewidth]{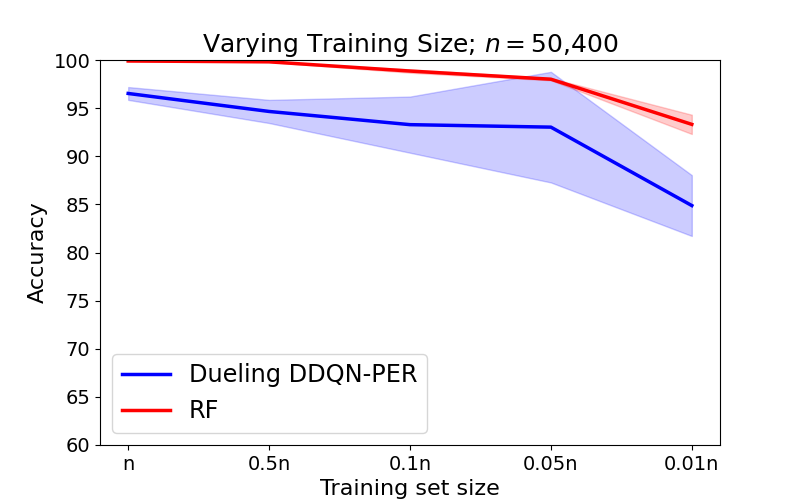}%
        \label{anem_varying_sizes}%
    }
    \caption{Accuracy of approaches with varying levels of missingness, noisiness and train set size for the \emph{anemia} dataset. \protect\subref{anem_missingness} shows the mean accuracy of the models at different missingness levels; \protect\subref{anem_noisiness} shows the mean accuracy of the models at different noisiness levels; \protect\subref{anem_missingness_noisiness} shows the mean accuracy of the models at a constant noisiness level (0.2) and different missingness levels; \protect\subref{anem_varying_sizes} shows the mean accuracy and the 95\% confidence interval of the models based on the size of the train set. The y-axis of all the plots starts at 60 to make the performance difference clearer.}
    \label{fig:anem_results_plots}
\end{figure*}

\begin{figure*}[!ht] 
    \centering
    \subfloat[][]{\includegraphics[width=0.45\textwidth]{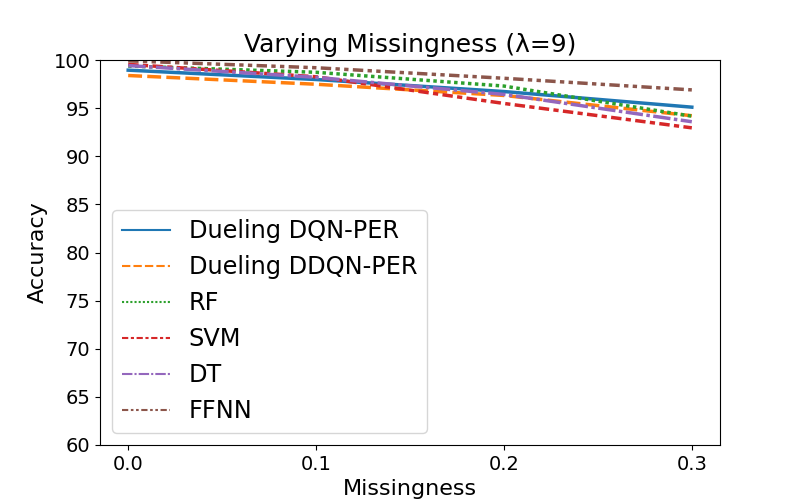}\label{lupus_missingness}} 
    \subfloat[][]{\includegraphics[width=0.45\textwidth]{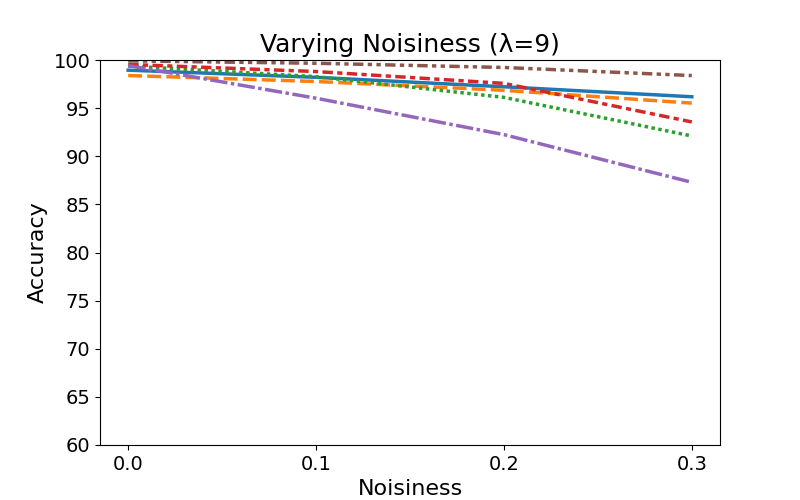}\label{lupus_noisiness}} \\
    \subfloat[][]{\includegraphics[width=0.45\textwidth]{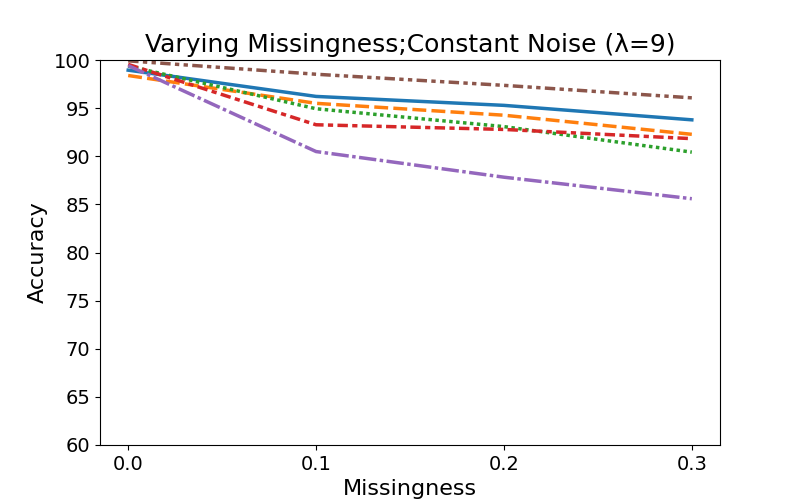}\label{lupus_missingness_noisiness}}
    \subfloat[][]{\includegraphics[width=0.45\textwidth]{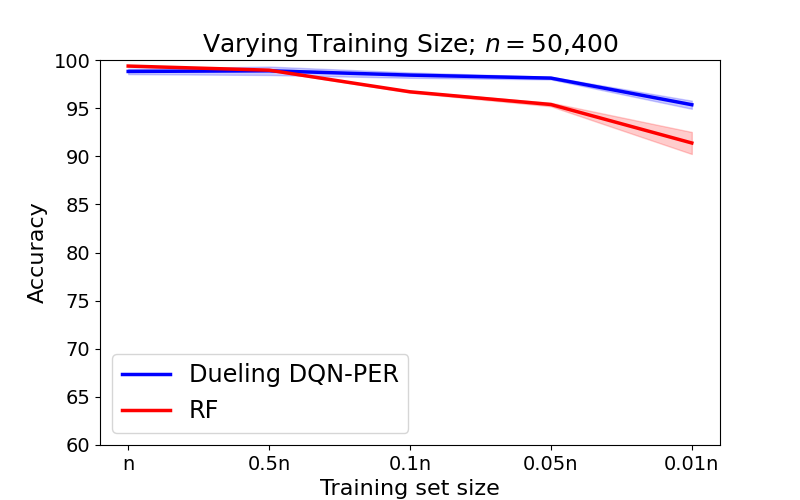}\label{lupus_varying_sizes}}
    \caption{Accuracy of approaches with varying levels of missingness, noisiness, and train set size, with the \emph{lupus} dataset. The graphs show the mean accuracy of the models at different \protect\subref{lupus_missingness} missingness levels; \protect\subref{lupus_noisiness} noisiness levels; \protect\subref{lupus_missingness_noisiness} missingness levels at a constant noisiness level (0.2). \protect\subref{lupus_varying_sizes} shows the mean accuracy and the 95\% confidence interval of the models as a function of the size of the train set. The y-axis of all the plots starts at 60 to make the performance difference clearer.}
    \label{fig:lupus_results_plots}
\end{figure*}

% \subsubsection{Anemia use case}
Figures ~\subref*{anem_missingness} and Figure~\subref*{lupus_missingness} depict the mean accuracy as the level of missing data in the \emph{anemia} and \emph{lupus} training datasets changes respectively. In particular, for lupus, before model training, the missing values were imputed using KNN imputer \cite{troyanskaya2001missing} with the number of neighbors set to one. All models exhibited a decline in performance, with SVM demonstrating the speediest deterioration in both use cases. The DT and FFNN had the best performance for \emph{anemia} and \emph{lupus} respectively.

Figures ~\subref*{anem_noisiness} and ~\subref*{lupus_noisiness} show the mean accuracy as the level of noise in both training sets increases. For anemia, SVM showed the sharpest decline, while the tree-based models (DT and RF) still had the best performance. On the other hand, for lupus, the FFNN had the best performance, with the DQN models outperforming the other models.  

Figures ~\subref*{anem_missingness_noisiness} and \subref*{lupus_missingness_noisiness} depict the varying mean accuracy as the level of missingness increases but with a fixed level of noise of 0.2. Here, again, for the \emph{lupus} use case, the missing values were imputed as described earlier. The DQN models has a better/steadier performance than all the models for anemia, while they had a slightly lower performance than the FFNN for lupus, but had a better performance than the rest of the models. In the \textit{lupus} use case, the tree-based models (DT and RF) had the worst performance, while SVM nad the worst for the \textit{anemia }use case followed by DT.

\subsection{Varying the size of train sets}
In order to understand whether this method can be applied to the diagnosis of conditions where data may be scarce, such as with rare diseases, we conducted experiments with train sets of different sizes. For each size, a fraction of the train set was randomly selected and removed, leaving the remaining data to form the new set. Runs were repeated five times for each size of the train set, whereby a different subset was removed each time. The Dueling DDQN-PER model was used for these experiments and for each run, the same number of timesteps was used. Figures~\subref*{anem_varying_sizes} and ~\subref*{lupus_varying_sizes} show the mean accuracy for each train set size for the two use cases. The shaded region represents the 95\% confidence interval. The RF classifier is also depicted in the figure for comparison purposes and it was chosen for its stability and high performance. 
As depicted, the DQN model was able to learn the clinical pathways and diagnose the patients. However, for all train set sizes, its performance was lower than the RF for the \emph{anemia} use case but was higher for the \emph{lupus} use case. Additionally, as the size of the train sets decreased, the model's performance became less stable.

\section{Discussion}\label{sec:discussion}
Our study illustrates that DRL models trained on EHRs are a promising solution to suggest optimal, individual, and step-by-step pathways to diagnosis.
% : they suggest taking a different action at each step until a diagnosis is reached.
In comparing our two best DQN models 
%\textit{i.e.}, Dueling DQN-PER and Dueling DDQN-PER, 
to state-of-the-art classifiers, while the latter performed extremely well on perfect data, the two DQN models performed comparably and sometimes better on imperfect data in both use cases (\textit{i.e.}, anemia and lupus), showing their robustness. %Additionally, for all the experiments, the DQN models outperformed SVM. 
Furthermore, using different training sizes, the DQN Models showed their capacity to learn diagnostic decision pathways associated with decent performance, even in the context of reasonably small-sized training datasets (\textit{e.g} about 600 patients). 

\subsection{Generated Pathways}
Besides quantitative evaluation, the pathways constructed by DQN models present two advantages. The fact that they are composed of progressive observations that lead to a diagnostic decision, makes them explainable; one may understand why a particular diagnosis is reached or why a new exam is ordered by looking at the sequence of features selected and their associated values, in a way that is similar to clinician reasoning. We note that the DT algorithms are also intrinsically explainable, but our results show that they are less robust to imperfect data. 

\begin{figure}[htbp]
    \centering
    \includegraphics[width=\linewidth]{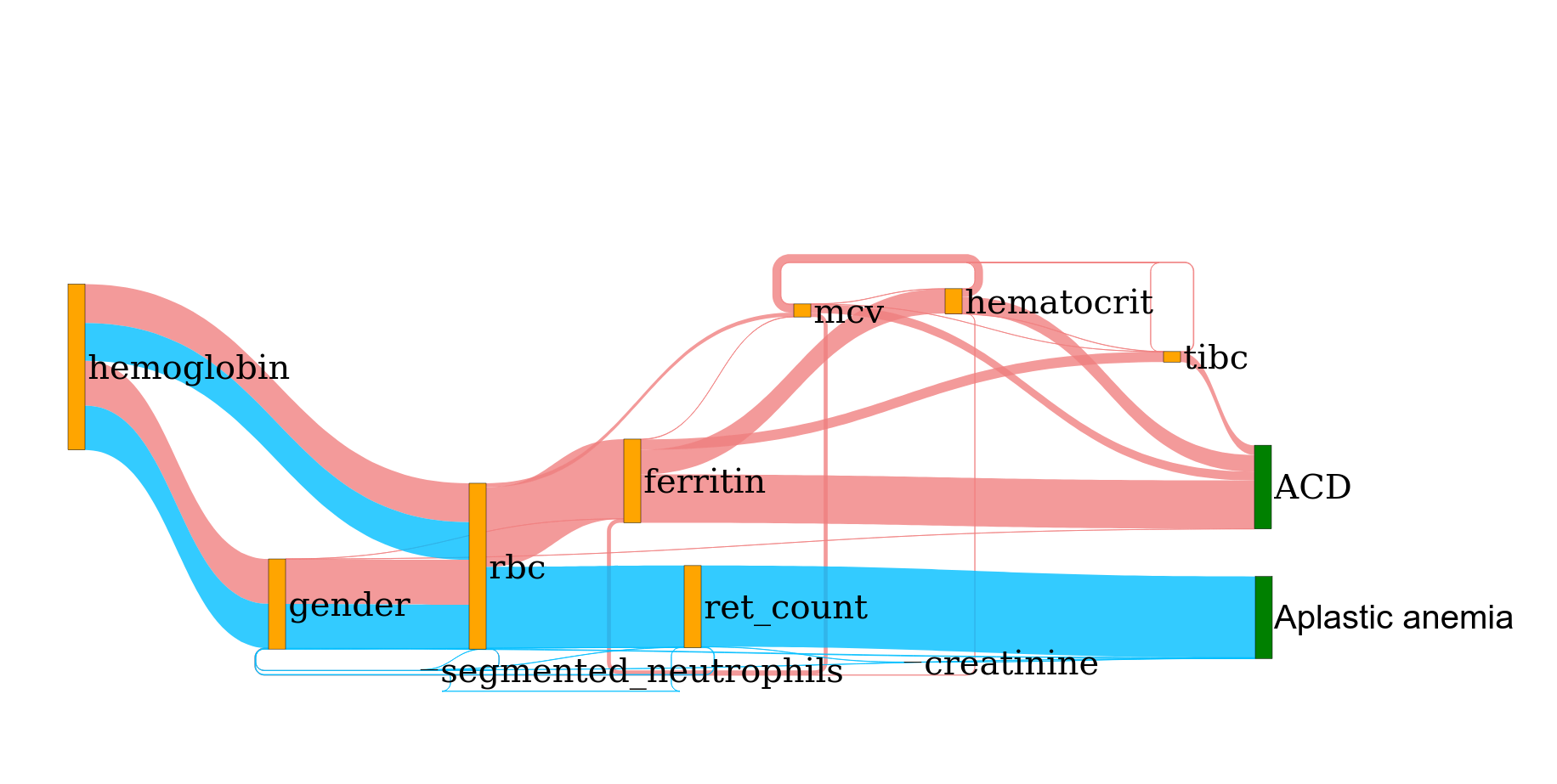}
    \caption{The diagnostic decision pathways for \emph{Anemia of Chronic Disease (ACD)} and \emph{Aplastic anemia} as learned by the agent.}
    \label{fig:anem_pathway2}
\end{figure}

\begin{figure}[htbp]
    \centering
    \includegraphics[width=\linewidth]{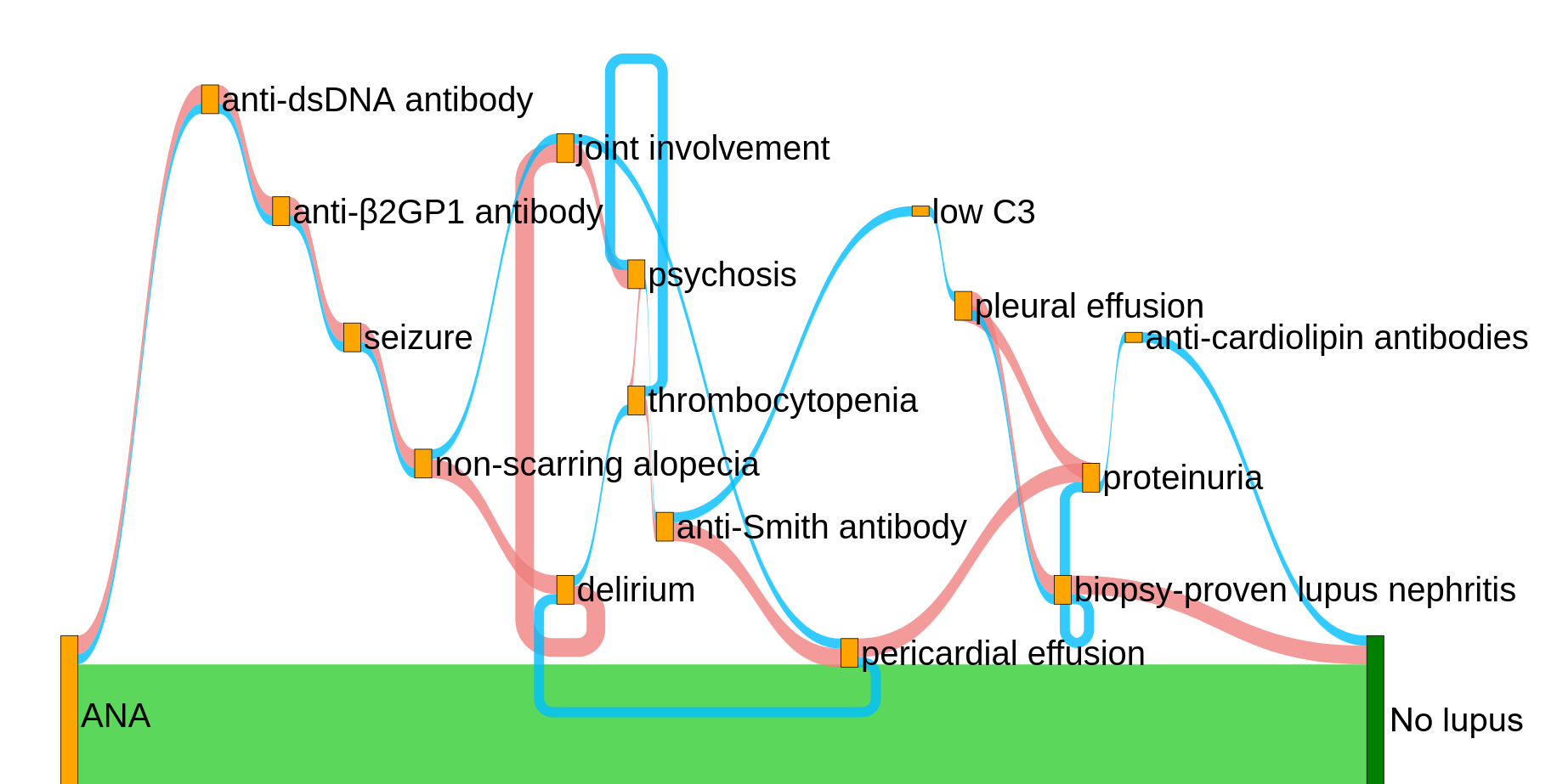}
    \caption{The three commonest diagnostic decision pathways for the \emph{No lupus} class generated by one of the models with the best wPAHM score. The model used to generate the pathways was the one with the wPAHM score closest to the mean wPAHM of the ten runs.}
    \label{fig:lupus_wpahm_no_lupus}
\end{figure}

In addition, the set of pathways generated by a test set can be aggregated to generate a data structure somewhat similar to diagnosis guidelines. 
% Figures \ref{fig:anem_pathway2} and \ref{fig:lupus_wpahm_no_lupus} illustrate such aggregations for the anemia and lupus use cases respectively. 
An illustration of such an aggregation is provided in Figure  
\ref{fig:anem_pathway2} in the form of a Sankey diagram. It shows the pathways learned by the model for the diagnosis of \emph{Anemia of Chronic Disease} and \emph{Aplastic anemia}, which are colored \emph{coral} and \emph{blue} respectively. A side-by-side illustration of some of the learned pathways and a branch of the decision tree used to label the dataset is also shown in Figure \ref{fig:side_by_side}.
Regarding the lupus use case, Figure \ref{fig:lupus_wpahm_no_lupus} illustrates the possible aggregations, showcasing the combination of the three commonest pathways for the \emph{No lupus} class. These pathways were generated by one of our experiments: the dueling DDQN model chosen based on the best wPAHM score, with $\lambda$ set to 9. More pathways are shown in Figures
% \cref{fig:lupus_wpahm_lupus, fig:lupus_acc_no_lupus, fig:lupus_acc_lupus} 
\ref{fig:lupus_wpahm_lupus}, \ref{fig:lupus_acc_no_lupus} and \ref{fig:lupus_acc_lupus} in \ref{apd:lupus}.

Other aggregations of pathways can be visualized and interacted with at \url{https://lilly-muyama.github.io/}. These provide visualization facilities and additional information about the support of nodes and edges, as well as descriptive statistics (\textit{e.g.}, mean, max, min values) of the various features.

% \begin{figure*}[!ht] 
%     \centering
%     % \subfloat{\includegraphics[width=0.45\textwidth]{images/pathways_lupus_acc_sized.png}\label{lupus_acc}} 
%     \subfloat{\includegraphics[width=0.45\textwidth]{images/pathways_lupus_wpahm_sized.png}\label{lupus_wpahm}}
%     \subfloat{\includegraphics[width=0.45\textwidth]{images/pathways_no_lupus_wpahm_sized2.png}\label{no_lupus_wpahm}}
%     \caption{The three commonest pathways for the \protect\subref{lupus_wpahm} Lupus and \protect\subref{no_lupus_wpahm} No lupus classes generated by one of the models with the best wPAHM scores. The model used to generate the pathways was the one with a wPAHM score closest to the mean wPAHM of the ten runs.}
%     \label{fig:lupus_wpahm_pathways}
% \end{figure*}

\subsection{DQN Performance}

Dueling DQN and Double DQN both improve the accuracy and stability of the standard DQN as observed in Table \ref{tab:dqn_results}. 
% However, in the anemia use case, Dueling DDQN exhibited the lowest performance of all the algorithms and showed a wide variability in its results, which was not the case in the lupus use case. Therefore, we hypothesize that this might be due to the nature of the data itself. 
%This may be because both Double DQN and Dueling DQN were created to solve the same problem: the overestimation of Q-values. 
% Using both techniques at the same time may not lead to better performance as long as the Q-value is still estimated using the same technique as in Double DQN. Additionally, using both Double DQN and Dueling DQN at the same time makes the model more complex, which may result in further instability. However, this was not replicated in the lupus use case.
Additionally, note that for each of the four algorithms discussed above, 
combining them with Prioritized Experience Replay resulted in 
better results both in terms of mean and SD of the accuracy. This is explained by the fact that in DQN, Double DQN, and Dueling DQN, during training, the experiences are sampled uniformly from the replay buffer. However, PER prioritizes these experiences such that those with a higher priority are sampled more often. In a task such as ours, this type of DQN favors transitions with a non-zero reward, which includes the diagnosis actions, on which the accuracy is primarily based. Sampling these actions more frequently leads the model to learn better and faster about them. 

% \subsection{Quality of anemia diagnoses}
\subsection{Quality of diagnoses}

For the \textbf{anemia} use case, regarding the comparison with SOTA classifiers, since we built our dataset using a decision tree, it was expected that the tree-based classifiers would perform extremely well. The performance of the two DQN Models was slightly lower than those of the tree-based methods and the neural network, while SVM had slightly lower performance.

Table \ref{tab:classification_report} displays the classification report of one particular experiment with the Dueling DQN-PER model, with an accuracy of 97.19, which is the closest accuracy to the mean. The report shows that most of the anemia classes exhibited a decent performance. However, while the \emph{No anemia} class had the best recall with a perfect score of 1, it also had the lowest precision at 0.92, meaning that several instances were being diagnosed as \emph{No anemia}, whereas they should not, based on their laboratory test results. Upon further inspection, we noted that all the misdiagnosed instances had values near the threshold value. In the \emph{No anemia} misclassifications, all the hemoglobin levels were close to the 13 and 12 g/dl thresholds for men and women, respectively (see Figure \ref{fig:tree}). In particular, the hemoglobin of men had a mean of 12.98 and a standard devisation of 0.01, while the one of women is 11.93 with SD=0.04. However, it should be noted that from a clinical point of view, hemoglobin levels below but close to the normal threshold usually do not lead to an anemia diagnosis. % into account since no specific therapeutic action is required.  
%Figure \ref{} shows the diagnosis pathways followed for the patients misdiagnosed with \emph{No anemia} 
In this regard, the decision tree of a guideline may lead to more false negatives than our approach. 
Similarly, looking at \emph{Hemolytic anemia}, which had the lowest recall score at 0.94 (along with \emph{Aplastic anemia}), 35 of its instances were diagnosed with \emph{Vitamin B12/Folate deficiency anemia}. These instances had a mean MCV value of 99.77 (SD=0.17), thus close to the threshold of 100. The 11 other instances diagnosed with \emph{Anemia of chronic disease} had a mean MCV level of 80.20 (SD=0.15), close to the threshold of 80. And the other 30 instances with an \emph{Inconclusive diagnosis} all had values near the thresholds, but missing values for other features in the pathway leading to an inconclusive diagnosis. 

\begin{table}[h]
  \fontsize{8pt}{8pt}\selectfont
  \centering
   % \scriptsize
  \begin{tabular}{c|c|c|c}
     %\hline
        \textbf{Class Name (support)} & \textbf{Precision} & \textbf{Recall} & \textbf{F1-Score} \\
        %& & &\\
        \hline
        \textbf{Inconclusive} & \multirow{2}{*}{0.94} & \multirow{2}{*}{0.97} & \multirow{2}{*}{0.95} \\
        \textbf{diagnosis (1344)}& & \\
        \hline
        \textbf{Aplastic} & \multirow{2}{*}{1.00} & \multirow{2}{*}{0.94} & \multirow{2}{*}{0.97} \\
        \textbf{anemia (1806)}& & \\
        \hline
        \textbf{Hemolytic} & \multirow{2}{*}{1.00} & \multirow{2}{*}{0.94} & \multirow{2}{*}{0.97} \\
        \textbf{anemia (1805)}& & \\
        \hline
        \textbf{Iron deficiency} & \multirow{2}{*}{0.98} & \multirow{2}{*}{0.98} & \multirow{2}{*}{0.98} \\
        \textbf{anemia (1679)}& & \\
        \hline
        \textbf{Anemia of chronic} & \multirow{2}{*}{0.99} & \multirow{2}{*}{0.97} & \multirow{2}{*}{0.98} \\
        \textbf{disease (1772)}& & \\
        \hline
        \textbf{Unspecified} & \multirow{2}{*}{1.00} & \multirow{2}{*}{0.98} & \multirow{2}{*}{0.99} \\
        \textbf{anemia (1793)}& & \\
        \hline
        \textbf{Vitamin B12/Folate} & \multirow{2}{*}{0.96} & \multirow{2}{*}{0.98} & \multirow{2}{*}{0.97} \\
        \textbf{defic. anemia (1801)}& & \\
        \hline
        \multirow{2}{*}{\textbf{No anemia (2000)}} & \multirow{2}{*}{0.92} & \multirow{2}{*}{1.00} & \multirow{2}{*}{0.96}\\
  \end{tabular}
  \caption{Classification report showing the detailed performance of the Dueling DQN-PER model. Classes of anemia diagnosis are reported with their respective support.}
  \label{tab:classification_report}
\end{table}

% Figure ~\subref*{lupus_wpahm_cm} shows the confusion matrix of the model that was used to generate the pathways in Figure \ref{fig:lupus_wpahm_pathways}. 
Regarding the \textbf{lupus} use case, for a positive SLE diagnosis to be made, the patient has to have a weighted criteria score of at least 10 (See \cite{aringer20192019} for details about how this total is calculated).
However, as reflected by the metrics, there were a few misdiagnosed episodes. For instance, the model whose pathways are presented in Figures \ref{fig:lupus_wpahm_no_lupus} and \ref{fig:lupus_wpahm_lupus}, 
% and ~\subref*{lupus_wpahm_cm}, 
had an accuracy and a wPAHM score of 96.57 and 91.65 respectively. 
However, on closer examination, it was found that for this particular model,
% As shown by its confusion matrix in Figure ~\subref*{lupus_wpahm_cm}, 
there were 130 \emph{Lupus} cases that were misdiagnosed as \emph{No lupus}. Subsequent analysis showed that this was primarily due to the pathways being terminated earlier than they should have. That is to say, some of the features, which were positive and would have led to a weighted criteria score greater than the threshold (10), were not tested. For example, the commonest pathway in these misdiagnosed episodes has a length of 15. Although these features are all key (the pathway still has an accuracy of 97.73\%), there are cases where the remaining nine features had values that would have changed the diagnosis. 

% \begin{figure}
%     \centering
%     \includegraphics[width=0.5\linewidth]{images/wpahm_21_confusion_matrix.png}
%     \caption{Confusion Matrix for dueling DDQN-PER model selected based on its wPAHM score.}
%     \label{fig:lupus_wpahm_cm}
% \end{figure}

% \begin{figure}[!ht] 
%     \centering
%     \subfloat[][]{\includegraphics[width=0.45\textwidth]{images/wpahm_21_confusion_matrix.png}\label{lupus_wpahm_cm}} 
%     \subfloat[][]{\includegraphics[width=0.45\textwidth]{images/acc_0_confusion_matrix.png}\label{lupus_acc_cm}} 
%     \caption{Confusion matrices for dueling DDQN-PER models selected based on their \protect\subref{lupus_wpahm_cm} wPAHM and \protect\subref{lupus_acc_cm} accuracy scores.}
%     \label{fig:lupus_cm}
% \end{figure}

Moreover, for the patients misdiagnosed as \emph{Lupus} instead of \emph{No lupus}, the model made the diagnosis too early when the weighted criteria score was near the threshold. For example, for this particular model, the commonest pathway for these misdiagnosed episodes was ``ANA=1 --> Anti-dsDNA antibody=1 --> Lupus''. The weighted criteria score of this pathway is less than 10 and yet a Lupus diagnosis was given. While the presence of anti-dsDNA antibody has one of the highest weights, it is not sufficient for a lupus diagnosis on its own according to \cite{aringer20192019}. Additionally, the second commonest pathway for these misdiagnosed episodes had a length of 23, missing only two features. On examining the features that were not selected by the model, it was revealed that one (renal biopsy-proven lupus nephritis) had the lowest penalty weight, while the other (acute pericarditis) had the lowest prevalence in the dataset.

It should be noted that the above error analysis concerns a model based on the best wPAHM score. The model with the best accuracy score had longer pathways, hence fewer misdiagnosed episodes.
% , as shown by the confusion matrix in Figure~\subref*{lupus_acc_cm}. 
Some of its pathways are shown in Figures \ref{fig:lupus_acc_no_lupus} and \ref{fig:lupus_acc_lupus} in \ref{apd:lupus}.

\subsection{Quality of lupus pathways}
\begin{figure}
    \centering
    \includegraphics[scale=0.5]{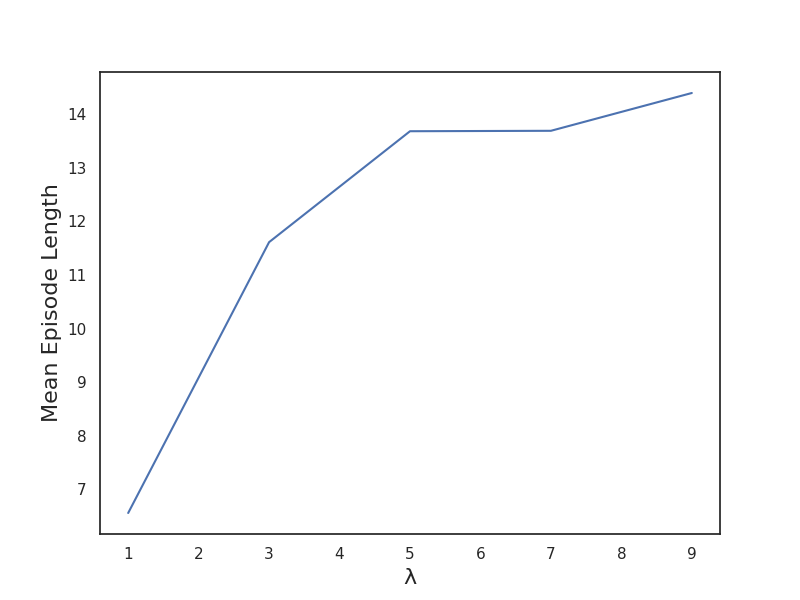}
    \caption{Mean episode length \textit{vs.} $\lambda$ for the Dueling DDQN-PER model.}
    \label{fig:lambda_vs_episode_length}
\end{figure}

When the value of $\lambda$ increases, the penalty for selecting the various actions decreases, hence the pathways become longer. While this leads to a higher diagnosis accuracy because the agent has more information to make the diagnosis, it also leads to a reduction in the quality of the diagnosis pathways, since actions that are not fully necessary may be taken. On the other hand, if $\lambda$ is too low, this leads to shorter pathways leading to a diagnosis being given before all the necessary actions have been conducted. 
Figure \ref{fig:lambda_vs_episode_length} shows the effect of the value of $\lambda$ has on the mean episode length for the dueling DDQN-PER model.

% For instance, comparing the performance of the Dueling DQN-PER models whose accuracy is closest to the mean accuracy for $\lambda=3$ and $\lambda=9$, we see that when $\lambda=9$, the model has an accuracy of 98.736 and a PAHM score of 22.403, while when $\lambda=3$, the model has a reduced accuracy of 97.171 but an increased PAHM score of 39.596 

% For instance, comparing the performance of the Dueling DDQN-PER models whose accuracy is closest to the mean accuracy for $\lambda=3$ and $\lambda=9$, we see that when $\lambda=9$, the model has an accuracy of 98.321, a PAHM score of 67.993, a wPAHM score of 90.269, and an average episode length of 12.556, while when $\lambda=3$, the model has a reduced accuracy of 95.964, a PAHM score of 63.709, a wPAHM score of 87.140, and an average episode length of 12.941. 

Additionally, when comparing the performance of the Dueling DDQN-PER models based on a single goal metric, \textit{i.e.}, by looking at the models with the best accuracy or wPAHM as shown in Table \ref{tab:metrics_results}, we see that with accuracy, the model has a lower average pathway score, a lower wPAHM score and a longer mean episode length (\textit{i.e.}, longer pathways), while with wPAHM as goal metric, we observe a lower accuracy, but a higher average pathway score and shorter episode lengths.

\subsection{DQN Model Robustness}

The DQN models show robustness to both noise and missing data, maintaining a consistent performance across various levels of increasing noise and missingness in both use cases. We believe that this robustness can be attributed to the reliance of the DQN on neural networks, which have been shown to be robust to noise in many scenarios \cite{goodfellow2016deep}. This is because adding noise to the training input can act as a form of regularization by preventing overfitting and leading to better generalization of the neural network. Likewise, neural networks have also demonstrated their capability to handle missing data as they are able to learn the underlying representation of the data even with missing values. 

For example, in the anemia use case, we remarked that introducing noise and missingness led to %significant 
decreased performance of the \emph{Inconclusive diagnosis} and \emph{Iron deficiency anemia} classes (results not shown). 
This is first attributed to the fact that some of the instances with noise or missingness are being diagnosed with \emph{Inconclusive diagnosis} instead of their anemia class, which is to be expected. In the real world, clinicians may need to use their own judgment in such cases. 
Second, regarding \emph{Iron Deficiency Anemia} misclassifications, those are almost all \emph{Anemia of Chronic Disease} instances that were falsely attributed to \emph{Iron Deficiency Anemia}, which are two classes difficult to discriminate, as illustrated by their location on the same branch of the decision tree (see Figure \ref{fig:tree}). 

We also note that in the \emph{lupus} use case, the Feed-Forward neural network has the best performance in all the various scenarios that were tested, followed closely by the DQN models, which performed better than all the tree-based models and SVM in all the scenarios with imperfect data. This can be attributed to the fact that the anemia dataset was labeled using a decision tree hence the very good performance of the tree-based models. However, since this is not the case with the lupus dataset, the neural network-based models performed better for that use case.

We also observe that while the DT performs very well in both use cases on the perfect data, its performance decreases when both noise and missing data are present, becoming significantly lower than that of the other models. This can be attributed to decision tree sensitivity to noise and missing data. In addition, decision trees are prone to overfitting, such that a slight change in the training dataset may lead to a different tree altogether. Alternatively, Random Forest builds multiple decision trees during their training process and is therefore less likely to overfit than the DT, making it more impervious to noise and missing data.

Finally, using the same hyperparameters for the model, the DQN model was able to interestingly perform consistently as the training set size gradually decreased. The reason for this is the same hyperparameters were used for the experiments therefore a smaller dataset just meant that the same experiences were sampled more often. However, there is still a loss in the information gained from the data, and therefore the smaller the dataset, the lower the accuracy and the more varied the results, as shown in Figures ~\subref*{anem_varying_sizes} and ~\subref*{lupus_varying_sizes}. In the anemia use case, the tree-based model (RF) performed better than the DQN Model, while the DQN model performed better than RF in the lupus use case. Again, this can be attributed to the different way diagnoses are made (\textit{i.e.}, decision tree- \textit{vs.} weighted criteria-based approach) in the two use cases.

Since we experimented on two use cases with different diagnosis processes, \textit{i.e.} decision tree and weighted criteria system, and we were able to achieve a good performance with both, we are therefore confident that our approach can generalize to other diagnoses that follow a similar diagnostic process as that of anemia or lupus. 
% However, it should be noted that the methodology may have to be adapted to the specific use case as was done in this study.

\subsection{Differences in use cases}

For exploration and experimentation purposes, we handled missing data differently in the two use cases. For lupus, we used KNN imputation. However, this was not considered necessary for the anemia use case with the presence of the \emph{inconclusive diagnosis}. Both methods ensured a diagnosis was given by the model even in the presence of missing data and we achieved good results with both methods making them both relevant to this work.

Additionally, for the reward function, we chose not to include a step penalty in the reward function of the anemia use case. Despite that, the model was able to learn the relevant features to test for each anemia class. However, because the diagnostic decision for lupus is more complex, it was relevant to add a penalty for each step taken by the agent in order to avoid overlong pathways.

Furthermore, we only applied penalty weights and the additional performance metrics (average pathway score, wPAHM score) to the lupus use case and not anemia because the lupus features were more and extremely diverse. All the anemia features were laboratory test results derived from a blood sample, while lupus features comprised physical symptoms such as fever, invasive procedures such as a renal biopsy, laboratory tests such as low C3, \textit{etc}.

Finally, we only added a maximum length constraint to the anemia dataset. 
This was dependent on the use case. 
Since the anemia diagnosis is reached via a decision tree, a maximum length could be interpreted as the maximum depth of the decision tree. Also, the anemia dataset included extra features not related to the anemia diagnosis. Therefore, in no case in the anemia use case is a pathway with all the features included needed to make a diagnosis. In contrast, for lupus, all the features in the dataset are relevant for the diagnosis, and in some cases, depending on the feature values, the agent may need to select all the features in the dataset before reaching a diagnosis. 
That said, experiments without the maximum length constraint for the anemia use case showed that the models were still able to learn the pathways with similar pathway lengths. Additionally, experiments with the lupus use case with the maximum length set to the number of features plus one (for the diagnosis) exhibited similar performance to the experiments without the maximum length whose results are shown in this work.

It is therefore important to note that our approach was adapted to each use case, guided by empirical results, to obtain optimal performance. Therefore, similar adjustments may be required for other use cases to ensure its effectiveness.

\subsection{Limitations}
One significant limitation of this study is the absence of experiments on a real-world dataset to evaluate our method. While the results from the synthetic datasets are very encouraging, and while we endeavor to reproduce imperfections of EHR data in our study, it is paramount that we test on real-world data. 
Additionally, the features that make up the pathways need to be available in the EHRs, which is not always the case.
Furthermore, the synthetic datasets used are not longitudinal. This initial study with anemia and lupus aimed to show the viability of the proposed method. Real-world EHR datasets often have a temporal dimension which will provide more insights into the progression of the pathways. With the availability of such data, we will be able to compare our approach with models capable of handling sequential data, such as Recurrent Neural Networks. 

Moreover, training the DRL agents took a significantly longer time and used considerably more computing resources than the SOTA classifiers as shown in Tables \ref{tab:lupus_resources} and \ref{tab:anem_resources} in the Appendices. Also, the DQN model has many hyperparameters that would need to be further optimized. However, it should be noted that after model training, generating a diagnosis pathway for a test instance is trivial since the policy has already been learned by the model (see Tables \ref{tab:lupus_resources} and \ref{tab:anem_resources} in the Appendices). 

Furthermore, there were slight differences in the methodology and performance evaluation strategy used in the two use cases. 
% \textit{handling missing data}. 
While we thought it was important to illustrate the results with these differences for research purposes, it should also be noted that they may account for some of the minor differences in the performance between the two use cases. 

% Finally, we did not evaluate Large Language Model-based methods, which have shown promising results on other tasks. We will consider them in future work.

\section{Conclusion}\label{sec:conclusion}

In this work, we adapted the RL framework to propose step-by-step guidance toward clinical diagnosis from synthetic EHRs. We particularly demonstrated the capacity of DRL to produce accurate diagnosis in the context of imperfect data (\textit{i.e.}, with noise, missingness), and variable dataset sizes; and an explainable diagnosis with the generation of decision pathways. 

Our approach with DQN is suitable as it has a comparable performance with the SOTA classifiers with the added advantage of constructing personalized pathways for each patient. 
These pathways have the potential to guide the diagnosis of a new patient or to be aggregated to summarize possible pathways for an uncommon patient population that might not be covered by guidelines in place.
In this manner, this approach has the potential to suggest enrichments for clinical guidelines, learned from practice data. %Additionally, we illustrated the robustness of our model to noise and missing values.
In future work, we aim to evaluate our approach on real-world EHRs, and to extend the type of considered input to multimodal data. Additionally, we will also consider the use of Large Language Model-based methods.

\section*{CRediT authorship contribution statement}
\textbf{Lillian Muyama:} Data Curation, Methodology, Software, Investigation, Validation, Visualization, Writing - original draft, Writing - Review \& Editing.
\textbf{Antoine Neuraz:} Conceptualization, Methodology, Validation, Resources, Writing - Review \& Editing, Supervision, Project Administration, Funding acquisition.
\textbf{Adrien Coulet:} Conceptualization, Methodology, Validation, Resources, Writing - Review \& Editing, Supervision, Project Administration, Funding acquisition.

\section*{Declaration of Competing Interests}
The authors declare that they have no known competing financial interests or personal relationships that could have appeared to influence the work reported in this paper.

\section*{Acknowledgements}\label{sec:acknowledgement}
This work is supported by the Inria CORDI-S Ph.D. program, and benefited from a government grant managed by the Agence Nationale de la Recherche under the France 2030 program, reference ANR-22-PESN-0007 ShareFAIR, and ANR-22-PESN-0008 NEUROVASC.

% \end{comment}

%% If you have bibdatabase file and want bibtex to generate the
%% bibitems, please use
%%
\newpage
\bibliographystyle{elsarticle-num} 
\bibliography{bibliography}

\begin{thebibliography}{10}
\expandafter\ifx\csname url\endcsname\relax
  \def\url#1{\texttt{#1}}\fi
\expandafter\ifx\csname urlprefix\endcsname\relax\def\urlprefix{URL }\fi
\expandafter\ifx\csname href\endcsname\relax
  \def\href#1#2{#2} \def\path#1{#1}\fi

\bibitem{field1990clinical}
M.~J. Field, K.~N. Lohr, et~al., Clinical practice guidelines, Directions for a new program (1990) 1990.

\bibitem{steinberg2011clinical}
E.~Steinberg, S.~Greenfield, D.~M. Wolman, M.~Mancher, R.~Graham, et~al., Clinical practice guidelines we can trust, National Academies Press, 2011.

\bibitem{adler2021}
J.~Adler-Milstein, J.~H. Chen, G.~Dhaliwal, {Next-Generation Artificial Intelligence for Diagnosis: From Predicting Diagnostic Labels to “Wayfinding”}, JAMA 326~(24) (2021) 2467--2468.

\bibitem{jensen2012}
P.~B. Jensen, L.~J. Jensen, S.~Brunak, Mining electronic health records: towards better research applications and clinical care, Nature Reviews Genetics 13~(6) (2012) 395--405.

\bibitem{lipton2015learning}
Z.~C. Lipton, D.~C. Kale, C.~Elkan, R.~Wetzel, Learning to diagnose with {LSTM} recurrent neural networks, arXiv preprint arXiv:1511.03677 (2015).

\bibitem{miotto2016}
R.~Miotto, L.~Li, B.~A. Kidd, J.~T. Dudley, Deep patient: an unsupervised representation to predict the future of patients from the electronic health records, Scientific Reports 6~(1) (2016) 1--10.

\bibitem{choi2016doctor}
E.~Choi, M.~T. Bahadori, A.~Schuetz, W.~F. Stewart, J.~Sun, {Doctor AI}: Predicting clinical events via recurrent neural networks, in: Machine learning for healthcare conference, PMLR, 2016, pp. 301--318.

\bibitem{bmj_anemia}
R.~Zaiden, Evaluation of anemia, BMJ Best Practice\url{https://bestpractice.bmj.com/topics/en-us/93/diagnosis-approach} Accessed: 2022-09-08 (2022).

\bibitem{zhang2015paving}
Y.~Zhang, R.~Padman, N.~Patel, Paving the {COW}path: Learning and visualizing clinical pathways from electronic health record data, Journal of Biomedical Informatics 58 (2015) 186--197.

\bibitem{perer2015mining}
A.~Perer, F.~Wang, J.~Hu, Mining and exploring care pathways from electronic medical records with visual analytics, Journal of Biomedical Informatics 56 (2015) 369--378.

\bibitem{baker2017process}
K.~Baker, E.~Dunwoodie, R.~G. Jones, A.~Newsham, O.~Johnson, C.~P. Price, J.~Wolstenholme, J.~Leal, P.~McGinley, C.~Twelves, et~al., Process mining routinely collected electronic health records to define real-life clinical pathways during chemotherapy, International Journal of Medical Informatics 103 (2017) 32--41.

\bibitem{huang2013latent}
Z.~Huang, X.~Lu, H.~Duan, Latent treatment pattern discovery for clinical processes, Journal of Medical Systems 37~(2) (2013) 1--10.

\bibitem{huang2014discovery}
Z.~Huang, W.~Dong, L.~Ji, C.~Gan, X.~Lu, H.~Duan, Discovery of clinical pathway patterns from event logs using probabilistic topic models, Journal of Bomedical Informatics 47 (2014) 39--57.

\bibitem{huang2018probabilistic}
Z.~Huang, Z.~Ge, W.~Dong, K.~He, H.~Duan, Probabilistic modeling personalized treatment pathways using electronic health records, Journal of Biomedical Informatics 86 (2018) 33--48.

\bibitem{lin2021personalized}
X.~Lin, Y.~Li, Y.~Xu, W.~Guo, W.~He, H.~Zhang, L.~Cui, C.~Miao, Personalized clinical pathway recommendation via attention based pre-training, in: 2021 IEEE International Conference on Bioinformatics and Biomedicine (BIBM), IEEE, 2021, pp. 980--987.

\bibitem{li2022electronic}
T.~Li, Z.~Wang, W.~Lu, Q.~Zhang, D.~Li, Electronic health records based reinforcement learning for treatment optimizing, Information Systems 104 (2022) 101878.

\bibitem{koshimizu2020prediction}
H.~Koshimizu, R.~Kojima, K.~Kario, Y.~Okuno, Prediction of blood pressure variability using deep neural networks, International Journal of Medical Informatics 136 (2020) 104067.

\bibitem{obaido2022interpretable}
G.~Obaido, B.~Ogbuokiri, T.~G. Swart, N.~Ayawei, S.~M. Kasongo, K.~Aruleba, I.~D. Mienye, I.~Aruleba, W.~Chukwu, F.~Osaye, et~al., An interpretable machine learning approach for {H}epatitis {B} diagnosis, Applied Sciences 12~(21) (2022) 11127.

\bibitem{zoabi2021machine}
Y.~Zoabi, S.~Deri-Rozov, N.~Shomron, Machine learning-based prediction of {COVID}-19 diagnosis based on symptoms, npj Digital Medicine 4~(1) (2021) 3.

\bibitem{kavya2021machine}
R.~Kavya, J.~Christopher, S.~Panda, Y.~B. Lazarus, Machine learning and {XAI} approaches for allergy diagnosis, Biomedical Signal Processing and Control 69 (2021) 102681.

\bibitem{li2021active}
Y.~Li, J.~Oliva, Active feature acquisition with generative surrogate models, in: International Conference on Machine Learning, PMLR, 2021, pp. 6450--6459.

\bibitem{janisch2019classification}
J.~Janisch, T.~Pevn{\`y}, V.~Lis{\`y}, Classification with costly features using deep reinforcement learning, in: Proceedings of the AAAI Conference on Artificial Intelligence, Vol.~33, 2019, pp. 3959--3966.

\bibitem{yu2023deep}
Z.~Yu, Y.~Li, J.~Kim, K.~Huang, Y.~Luo, M.~Wang, Deep reinforcement learning for cost-effective medical diagnosis, arXiv preprint arXiv:2302.10261 (2023).

\bibitem{tang2016inquire}
K.-F. Tang, H.-C. Kao, C.-N. Chou, E.~Y. Chang, Inquire and diagnose: Neural symptom checking ensemble using deep reinforcement learning, in: NIPS Workshop on Deep Reinforcement Learning, 2016.

\bibitem{wei2018task}
Z.~Wei, Q.~Liu, B.~Peng, H.~Tou, T.~Chen, X.-J. Huang, K.-F. Wong, X.~Dai, Task-oriented dialogue system for automatic diagnosis, in: Proceedings of the 56th Annual Meeting of the Association for Computational Linguistics (Volume 2: Short Papers), 2018, pp. 201--207.

\bibitem{kao2018context}
H.-C. Kao, K.-F. Tang, E.~Chang, Context-aware symptom checking for disease diagnosis using hierarchical reinforcement learning, in: Proceedings of the AAAI Conference on Artificial Intelligence, Vol.~32, 2018.

\bibitem{Littman2001markov}
M.~Littman, Markov decision processes, in: N.~J. Smelser, P.~B. Baltes (Eds.), International Encyclopedia of the Social \& Behavioral Sciences, Pergamon, Oxford, 2001, pp. 9240--9242.

\bibitem{sutton2018reinforcement}
R.~S. Sutton, A.~G. Barto, Reinforcement learning: An introduction, MIT press, 2018.

\bibitem{watkins1992q}
C.~J. Watkins, P.~Dayan, Q-learning, Machine learning 8~(3) (1992) 279--292.

\bibitem{mnih2015human}
V.~Mnih, K.~Kavukcuoglu, D.~Silver, A.~A. Rusu, J.~Veness, M.~G. Bellemare, A.~Graves, M.~Riedmiller, A.~K. Fidjeland, G.~Ostrovski, et~al., Human-level control through deep reinforcement learning, Nature 518~(7540) (2015) 529--533.

\bibitem{brockman2016openai}
G.~Brockman, V.~Cheung, L.~Pettersson, J.~Schneider, J.~Schulman, J.~Tang, W.~Zaremba, Open{AI} {G}ym, arXiv preprint arXiv:1606.01540 (2016).

\bibitem{stable-baselines}
A.~Hill, A.~Raffin, M.~Ernestus, A.~Gleave, A.~Kanervisto, R.~Traore, P.~Dhariwal, C.~Hesse, O.~Klimov, A.~Nichol, M.~Plappert, A.~Radford, J.~Schulman, S.~Sidor, Y.~Wu, Stable {B}aselines, \url{https://github.com/hill-a/stable-baselines} (2018).

\bibitem{aringer20192019}
M.~Aringer, K.~Costenbader, D.~Daikh, R.~Brinks, M.~Mosca, R.~Ramsey-Goldman, J.~S. Smolen, D.~Wofsy, D.~T. Boumpas, D.~L. Kamen, et~al., 2019 {E}uropean {L}eague {A}gainst {R}heumatism/{A}merican {C}ollege of {R}heumatology classification criteria for {S}ystemic {L}upus {E}rythematosus, Arthritis \& Rheumatology 71~(9) (2019) 1400--1412.

\bibitem{short2013iron}
M.~W. Short, J.~E. Domagalski, Iron deficiency anemia: evaluation and management, American Family Physician 87~(2) (2013) 98--104.

\bibitem{troyanskaya2001missing}
O.~Troyanskaya, M.~Cantor, G.~Sherlock, P.~Brown, T.~Hastie, R.~Tibshirani, D.~Botstein, R.~B. Altman, Missing value estimation methods for dna microarrays, Bioinformatics 17~(6) (2001) 520--525.

\bibitem{goodfellow2016deep}
I.~Goodfellow, Y.~Bengio, A.~Courville, Deep learning, MIT press, 2016.

\bibitem{van2016deep}
H.~Van~Hasselt, A.~Guez, D.~Silver, Deep reinforcement learning with double {Q}-learning, in: Proceedings of the AAAI conference on artificial intelligence, Vol.~30, 2016.

\bibitem{wang2016dueling}
Z.~Wang, T.~Schaul, M.~Hessel, H.~Hasselt, M.~Lanctot, N.~Freitas, Dueling network architectures for deep reinforcement learning, in: International conference on machine learning, PMLR, 2016, pp. 1995--2003.

\bibitem{schaul2015prioritized}
T.~Schaul, J.~Quan, I.~Antonoglou, D.~Silver, Prioritized experience replay, arXiv preprint arXiv:1511.05952 (2015).

\bibitem{ht_formula}
M.~Koperska, Hematocrit/{H}emoglobin ratio calculator, \url{https://www.omnicalculator.com/health/hct-hgb}, accessed: 2022-11-22 (2023).

\bibitem{tsat_formula}
D.~Nedea, Transferrin saturation calculator, \url{https://www.mdapp.co/transferrin-saturation-calculator-444/}, accessed: 2022-11-24 (2020).

\bibitem{rbc_formula}
F.~Naeim, P.~{Nagesh Rao}, S.~X. Song, W.~W. Grody, 61 - disorders of red blood cells—anemias, in: F.~Naeim, P.~{Nagesh Rao}, S.~X. Song, W.~W. Grody (Eds.), Atlas of Hematopathology, Academic Press, 2013, pp. 675--704.

\bibitem{timlin2018fevers}
H.~Timlin, A.~Syed, U.~Haque, B.~Adler, G.~Law, K.~Machireddy, R.~Manno, Fevers in adult lupus patients, Cureus 10~(1) (2018).

\bibitem{fayyaz2015haematological}
A.~Fayyaz, A.~Igoe, B.~T. Kurien, D.~Danda, J.~A. James, H.~A. Stafford, R.~H. Scofield, Haematological manifestations of lupus, Lupus Science \& Medicine 2~(1) (2015) e000078.

\bibitem{galanopoulos2017lupus}
N.~Galanopoulos, A.~Christoforidou, Z.~Bezirgiannidou, Lupus thrombocytopenia: pathogenesis and therapeutic implications, Mediterranean Journal of Rheumatology 28~(1) (2017) 20--26.

\bibitem{crawford2022refractory}
L.~R. Crawford, N.~Neparidze, Refractory autoimmune hemolytic anemia in a {S}ystemic {L}upus {E}rythematosus patient: A clinical case report, Clinical Case Reports 10~(3) (2022) e05583.

\bibitem{nayak2012psychosis}
R.~B. Nayak, G.~S. Bhogale, N.~M. Patil, S.~S. Chate, Psychosis in patients with {S}ystemic {L}upus {E}rythematosus, Indian Journal of Psychological Medicine 34~(1) (2012) 90--93.

\bibitem{desai2021recent}
K.~Desai, M.~Miteva, Recent insight on the management of {L}upus {E}rythematosus {A}lopecia, Clinical, Cosmetic and Investigational Dermatology (2021) 333--347.

\bibitem{kudsi2021prevalence}
M.~Kudsi, L.~D. Nahas, R.~Alsawah, A.~Hamsho, A.~Omar, The prevalence of oral mucosal lesions and related factors in {S}ystemic {L}upus {E}rythematosus patients, Arthritis Research \& Therapy 23 (2021) 1--5.

\bibitem{gronhagen2014cutaneous}
C.~M. Gr{\"o}nhagen, F.~Nyberg, Cutaneous lupus erythematosus: An update, Indian Dermatology Online Journal 5~(1) (2014) 7.

\bibitem{provost1994relationship}
T.~T. Provost, The relationship between discoid and {S}ystemic {L}upus {E}rythematosus, Archives of {D}ermatology 130~(10) (1994) 1308--1310.

\bibitem{yao2020clinical}
X.~Yao, M.~Abd~Hamid, A.~Sundaralingam, A.~Evans, R.~Karthikappallil, T.~Dong, N.~M. Rahman, N.~I. Kanellakis, Clinical perspective and practices on pleural effusions in chronic systemic inflammatory diseases, Breathe 16~(4) (2020).

\bibitem{almousa2022unusual}
S.~Almousa, H.~Wannous, K.~Khedr, H.~Qasem, Unusual case presentation of {S}ystemic {L}upus {E}rythematosus in a young woman, Rheumato 2~(4) (2022) 93--97.

\bibitem{narang2022acute}
V.~K. Narang, J.~Bowen, O.~Masarweh, S.~Burnette, M.~Valdez, L.~Moosavi, F.~Joolhar, T.~T. Win, Acute pericarditis leading to a diagnosis of {SLE}: a case series of 3 patients, Journal of Investigative Medicine High Impact Case Reports 10 (2022) 23247096221077832.

\bibitem{ceccarelli2017joint}
F.~Ceccarelli, C.~Perricone, E.~Cipriano, L.~Massaro, F.~Natalucci, G.~Capalbo, I.~Leccese, D.~Bogdanos, F.~R. Spinelli, C.~Alessandri, et~al., Joint involvement in {S}ystemic {L}upus {E}rythematosus: from pathogenesis to clinical assessment, in: Seminars in Arthritis and Rheumatism, Vol.~47, Elsevier, 2017, pp. 53--64.

\bibitem{nasim2022low}
N.~Wiegley, Low-grade proteinuria in patients with {S}ystemic {L}upus {E}rythematosus, Kidney News 14~(10/11) (2022) 63 -- 63.

\bibitem{hong2018systematic}
W.~Hong, Y.-L. Ren, J.~Chang, G.~Luo, S.~Ling-Yun, A systematic review and meta-analysis of prevalence of biopsy-proven lupus nephritis, Archives of {R}heumatology 33~(1) (2018) 17.

\bibitem{unlu2016clinical}
O.~{\"U}nl{\"u}, S.~Zuily, D.~Erkan, The clinical significance of antiphospholipid antibodies in {S}ystemic {L}upus {E}rythematosus, European Journal of Rheumatology 3~(2) (2016) 75.

\bibitem{dema2016autoantibodies}
B.~Dema, N.~Charles, Autoantibodies in {SLE}: specificities, isotypes and receptors, Antibodies 5~(1) (2016) 2.

\bibitem{ramos2004hypocomplementemia}
M.~Ramos-Casals, M.~Campoamor, A.~Chamorro, G.~Salvador, S.~Segura, J.~Botero, J.~Yag{\"u}e, R.~Cervera, M.~Ingelmo, J.~Font, Hypocomplementemia in {S}ystemic {L}upus {E}rythematosus and primary antiphospholipid syndrome: prevalence and clinical significance in 667 patients, Lupus 13~(10) (2004) 777--783.

\bibitem{fabrizio2015systemic}
C.~Fabrizio, C.~Fulvia, P.~Carlo, M.~Laura, M.~Elisa, M.~Francesca, S.~Francesca~Romana, T.~Simona, A.~Cristiano, V.~Guido, et~al., {S}ystemic {L}upus {E}rythematosus with and without anti-ds{DNA} antibodies: analysis from a large monocentric cohort, Mediators of Inflammation 2015 (2015).

\bibitem{arroyo2015clinical}
M.~Arroyo-{\'A}vila, Y.~Santiago-Casas, G.~McGwin, R.~S. Cantor, M.~Petri, R.~Ramsey-Goldman, J.~D. Reveille, R.~P. Kimberly, G.~S. Alarc{\'o}n, L.~M. Vil{\'a}, et~al., Clinical associations of anti-{S}mith antibodies in {PROFILE}: a multi-ethnic lupus cohort, Clinical Rheumatology 34 (2015) 1217--1223.

\end{thebibliography}

\appendix

\counterwithin{figure}{section}
\counterwithin{table}{section}

\renewcommand{\thetable}{\Alph{section}.\arabic{table}}
\renewcommand{\thefigure}{\Alph{section}.\arabic{figure}} 

\newpage
\section{Deep-Q Network (DQN)}\label{apd:general}
\subsection{DQN and its extensions}\label{apd:Q}

\textbf{Q-learning} \cite{watkins1992q} is an RL algorithm that outputs the best action to take in a given state based on the expected future reward of taking that action in that particular state. The expected future reward is named the Q-value of that state-action pair, %\textit{i.e.}
noted $Q(s,a)$. At each time step, the agent selects an action following a policy $\pi$, and the goal is to find the optimal policy $\pi^*$ that maximizes the reward function. 
%In Q-learning, Q-values are initialized to 0 and stored in a look-up table. As the agent interacts with the environment, 
During model training, the Q-values are updated using the Bellman Equation as follows:

\begin{equation}
\label{eqn:q_learning}
  Q_{t+1}(s_t, a_t) \leftarrow Q_{t}(s_t, a_t) + \alpha[r_{t+1} + \gamma\  \underset{a}{\max}\  Q_{t}(s_{t+1}, a) - Q_{t}(s_t, a_t)]
\end{equation}

where $\alpha$ is the learning rate and $\gamma$ is the discount factor that determines the importance of future reward relative to immediate reward.

% A look-up table is only feasible for problems with relatively small finite state and action spaces. 

In our use case, since the problem has a large state space, we propose to use a \textbf{Deep Q-Network} \cite{mnih2015human} which uses a neural network to approximate the Q-value function.
A DQN comprises two networks of similar architecture, \textit{i.e.}, the policy network, which interacts with the environment and learns the optimal policy, and the target network, which is used to define the target Q-value. At each time step, the policy network takes a state, $s_{t}$, as its input and outputs the Q-values for taking the different actions in that state. The weights of the target network are frozen and updated at a specified interval by copying the weights of the policy network. The DQN algorithm learns by minimizing the loss function shown in Equation \ref{eqn:dqn_loss_fn}, 
%which is the mean squared error between the target and the predicted Q-values,
where $\theta$ and $\theta^-$ represent the weights of the policy and target networks, respectively. Additionally, at each time step, a record of the model's interaction with the environment (known as an experience) is stored in a memory buffer in the form $(s_t, a_t, r_{t+1}, s_{t+1})$.

\begin{equation}
    \label{eqn:dqn_loss_fn}
    L(\theta) = \mathop{\mathbb{E}}[(r_{t+1} + \gamma\ \underset{a}{\max}\  Q(s_{t+1}, a, \theta^-) - Q(s_t, a_t, \theta))^2]
\end{equation}

In order to improve DQN stability and performance, several extensions of the DQN algorithm have been developed which we use in this paper and we briefly describe below:

\textbf{Double DQN (DDQN)} \cite{van2016deep}: 
The action is selected using the policy network, while the target network estimates the value of that action unlike in the standard DQN where the same network is used for both tasks. The loss function is thus modified as follows:

\begin{equation}
    \label{eqn:ddqn_loss_fn}
    L(\theta) = \mathop{\mathbb{E}}[(r_{t+1} + \gamma\ Q(s_{t+1}, \underset{a}{\arg\max}\ Q(s_{t+1}, a, \theta), \theta^-) - Q(s_t, a_t, \theta))^2]  
\end{equation}

\textbf{Dueling DQN} \cite{wang2016dueling}: The Q-value function is split into two parts: a value function $V(s)$ that provides the value for being in that state, and an advantage function $A(s,a)$ that gives the advantage of the action $a$ in the state $s$, as compared to the other actions. The two functions are then combined to get the Q values as shown in Equation \ref{eqn:dueling_dqn}.

\begin{equation}
\label{eqn:dueling_dqn}
    Q(s,a) = V(s) + (A(s,a) - \frac{1}{|\mathcal{A}|} \underset{a}\sum A(s,a))
\end{equation}

where $|\mathcal{A}|$ is the total number of possible actions in that state.

\textbf{Prioritized Experience Replay (PER)} \cite{schaul2015prioritized}: 
Each experience in the buffer is assigned a priority such that experiences that have a higher priority are sampled more often during training. 

\subsection{Model Hyperparameters}\label{apd:params}
Table \ref{tab:hyperparams} shows the values of the hyperparameters used for the DQN model in this study.

\begin{table}[!ht]
\centering
    \begin{tabular}{|l|l|}
    \hline
    \textbf{Hyperparameter} & \textbf{Value} \\ \hline
        Buffer size & 1000000 \\ 
        Learning rate & 0.0001 \\
        Target network update frequency & 10000 \\ 
        Learning starts & 50000 \\
        Final epsilon value & 0.05 \\
        Discount factor & 0.99 \\
        Train frequency & 4\\ \hline
    \end{tabular}
    \caption{DQN hyperparameter values.}
    \label{tab:hyperparams}
\end{table}

%=================================================================
%=================================================================
%=================================================================
%============appendix Dataset================
\newpage
\section{Dataset Construction}\label{apd:data}
%\section{Dataset}\label{sec:cohort}
To experiment with our method, we built two synthetic datasets, one for each use case.

\subsection{Anemia Dataset}

\subsubsection{Feature Inclusion}
The first step of our dataset construction was the definition of a set of 17 features, associated or not, with anemia diagnosis. 
The hemoglobin level is the primary feature that is considered to determine whether a patient has anemia, therefore was included. Since the normal level of hemoglobin varies between men and women, gender was included as well. We also included features from a decision tree of the standard guidelines for the diagnosis of anemia by Zaïden~\cite{bmj_anemia}: mean corpuscular volume (MCV), ferritin, reticulocyte count, segmented neutrophils and Total Iron Binding Capacity (TIBC).

Additionally, we included features that are not in the tree, but can be used to diagnose anemia, according to discussions with a domain expert: hematocrit, transferrin saturation (TSAT), red blood cells (RBC), serum iron, and folate. Interestingly for our study, three of these features (hematocrit, TSAT and RBC) can be derived from other features and are accordingly correlated with features from our initial selection (\textit{e.g.}, hemoglobin and hematocrit).
Furthermore, we added features that are not relevant to the diagnosis of anemia to observe their potential impact on the behavior of our model: creatinine, cholesterol, copper, ethanol, and glucose. 

\subsubsection{Dataset Construction}\label{subsec:anem_data_construction}
The second step of the dataset construction is the instantiation of features with values associated with one of 7 diagnosis classes: \emph{No anemia}, \emph{Vitamin B12/Folate deficiency anemia}, \emph{Unspecified anemia}, \emph{Anemia of chronic disease (ACD)}, \emph{Iron deficiency anemia (IDA)}, \emph{Hemolytic anemia}, and \emph{Aplastic anemia}. Each anemia class dataset was built based on the decision tree represented in Figure \ref{fig:tree}, which was manually constructed based on \cite{bmj_anemia,short2013iron}.

For each class, the values of the features were generated using a uniform probability distribution whose parameters (min and max values), were determined through manual reviewing of the medical literature and thresholds of the decision tree in Figure \ref{fig:tree}.  Features that are correlated with others were derived using known equations, that is, hematocrit \cite{ht_formula}, TSAT \cite{tsat_formula} and RBC \cite{rbc_formula}.
%If a feature was relevant to a particular diagnosis class, \textit{i.e.}, it exists in the branch of the tree that leads to that diagnosis class, the range and thus the distribution from which the value was generated was different from that used when the feature was not associated with that class. However, for ACD and IDA, TIBC values were generated dependent on the ferritin value such that the range used to generate the TIBC value was based on whether the ferritin value was greater than 100 or not. Additionally, for features that were not relevant to the diagnosis of anemia class (based on the tree in Figure \ref{fig:tree}), a fraction (chosen randomly between 0.1 and 0.7) of their values were replaced with missing values since this did not affect the diagnosis. 
10,000 instances were created for each of the diagnosis classes, which were in turn combined to create a single dataset of 70,000 instances. Next, an eighth diagnosis class, \emph{Inconclusive diagnosis} was created for cases when the model is unsure about the diagnosis of an instance. The \emph{Inconclusive diagnosis} instances were created out of the existing 70,000 instances by randomly selecting and removing 10\% of the non-missing values of each feature, except for hemoglobin, gender and MCV, which are necessary to the diagnosis of almost all the anemia classes. 
%This left some instances with insufficient data to reach a diagnosis using the decision tree. These were labeled as \emph{Inconclusive diagnosis}. 
% The final dataset, therefore, had 17 features and 8 classes. 
In  \ref{apd:anemia}, %\ref{sec:appendix_a}, 
the class distribution is illustrated in Figure \ref{fig:anem_class_distribution}; the ratio of observed \textit{vs.} missing values for each feature is shown in Figure \ref{fig:anem_missing_values}; an example of an instance from the dataset is provided Table \ref{tab:anem_sample}; and the summarized statistics of the dataset are shown in Table \ref{tab:anem_statistics}.

\subsubsection{Simulating Imperfect Data}
To compare the robustness of various approaches to imperfect data, we artificially introduced different levels of noise and missingness to our training data.

A percentage of the values of each feature, excluding hemoglobin and gender, were randomly replaced with missing values. Hemoglobin and gender were excepted because a patient's hemoglobin level is key to the diagnosis of anemia and the normal levels of hemoglobin vary between men and women. 

Since our original dataset perfectly follows a decision tree, we simulated noisiness using the following procedure: For each anemia class, except \emph{No anemia} and \emph{Inconclusive diagnosis}, a specified fraction of the values in each of the features in its branch of the tree in Figure \ref{fig:tree} were replaced by another value generated from a normal distribution, $N(\mu, \sigma^2)$, where the mean, $\mu$, is the threshold in a node with that feature in the tree, and the standard deviation, $\sigma$, was defined by us arbitrarily.

For example, using hemolytic anemia, the features in its tree branch, that is, the features used to diagnose a patient with it are \emph{hemoglobin}, \emph{MCV} and \emph{reticulocyte count} as shown in Figure \ref{fig:tree}. Therefore, for a noise level of 0.2, we replaced 20\% of the \textit{reticulocyte count} values using a normal distribution function $N(2, 0.2)$ where $\mu$ value is picked following the threshold in the decision tree. Similarly, for \textit{MCV}, 10\% of its feature values were replaced using a normal distribution $N(80, 2)$, while another 10\% was replaced using a normal distribution function $N(100, 2)$. We gradually increased the noise level fraction in order to analyze its effect on the performance of our model and the pathways it creates. Additionally, for all the noise levels, 10\% of the anemic instances were randomly labeled as \textit{No anemia}. 

We further created datasets with both noisy and missing data. Using a training dataset with a fixed noisiness level of 0.2 as described above, we added missing data at different levels to this dataset using the same procedure used to create the datasets with missing data.

\subsection{Lupus Dataset}
\subsubsection{Feature Inclusion}
To construct the dataset, we generated features based on the clinical practice guideline outlined in \cite{aringer20192019} for the classification of Systemic Lupus Erythematosus (SLE). In adherence to this guideline, we identified a total of 24 features, namely, antinuclear antibodies (ANA), fever, leukopenia, thrombocytopenia, autoimmune hemolysis, delirium, psychosis, seizure, non-scarring alopecia, oral ulcers, cutaneous lupus, pleural effusion, pericardial effusion, acute pericarditis, joint involvement, proteinuria, renal biopsy-proven lupus nephritis, anti-cardiolipin antibodies, anti-$\beta$2GP1 antibodies, lupus anticoagulant, low C3, low C4, anti-dsDNA antibody, and anti-Smith antibody.

This guideline describes a weighted score using these features to determine whether an individual has SLE or not. Consequently, we used the same criteria to label our synthetic instances, establishing two classes, \textit{i.e.}, \emph{Lupus} and \emph{No lupus}.

\subsubsection{Dataset Construction}
With reference to \cite{aringer20192019}, we consider the presence of ANA at a titer $\geq$ 1:80 on HEP-2 cells as an entry criterion (referred to as the ANA feature for clarity) for SLE diagnosis, and accordingly we constructed two datasets. In one, instances were labeled as 1 (True) for the ANA feature, indicating a positive ANA result; in the second dataset, instances were labeled as 0 (False), indicating a negative ANA result. Specifically, we generated 50,000 instances with a positive ANA and 20,000 instances with a negative ANA.

In both datasets, all features are binary indicating whether a specific feature is present (1) or absent (0) in a given instance; except for two features: cutaneous lupus and renal biopsy-proven lupus nephritis that are categorical, taking on one of four values (0 for False, 1 for subacute cutaneous lupus, 2 for acute cutaneous lupus, 3 for discoid lupus) for cutaneous lupus, and one of six values (0 for False, 1 for class I, 2 for class II, 3 for class III, 4 for class IV, 5 for class V) for renal biopsy-proven lupus nephritis, respectively.

In the positive ANA dataset, the prevalence of positive features was determined from literature sources (see Table \ref{tab:lupus_literature_fractions} in \ref{apd:lupus}). Conversely, for the negative ANA dataset, the prevalence of features was derived considering their relative importance in lupus diagnosis, as indicated by their weight in the diagnosis criteria. Features with larger weights were considered less prevalent in patients, following the logic that their rarity contributes to their significance in the diagnosis process.

The features were populated using a weighted probability distribution, taking into account the prevalence levels or weighted criteria as previously outlined. Subsequently, the two datasets were labeled based on the weighted criteria specified in \cite{aringer20192019}, then merged to create a unified dataset of 70,000 instances and 24 features, where 35,056 and 34,944 instances belonged to the \textit{No lupus} and \textit{Lupus} classes, respectively. 
%Figure \ref{fig:class_distribution} shows the class distribution of the dataset,
Table \ref{tab:lupus_data_instance} provides an example of an instance. Table \ref{tab:lupus_summary_statistics} presents descriptive statistics for the dataset.

\subsubsection{Simulating Imperfect Data}
A percentage of values for each feature, excluding ANA, were randomly replaced with missing values. ANA was excluded because, as mentioned earlier, it serves as the entry criterion for SLE diagnosis therefore without this feature, a patient should not be diagnosed with lupus according to \cite{aringer20192019}.

Additionally, we introduced noisiness into the dataset through the following method. For each level of noisiness, a specified fraction of instances in the dataset was randomly selected, and their labels were changed to the opposite class. For instance, at a noisiness level of 0.2, the diagnoses of 20\% of the instances in the training set were altered; those initially diagnosed with \textit{Lupus} were changed to \textit{No lupus}, and vice versa. Consequently, the training dataset no longer perfectly adhered to the diagnosis criteria outlined in \cite{aringer20192019}.

Furthermore, we generated datasets incorporating both noisy and missing data. Starting with a training dataset featuring a noisiness level fixed at 0.2, we introduced missing data at various levels using the same procedure employed in creating datasets with missing values. This approach allowed us to explore the impact of both noise and missing information on the dataset simultaneously.

%=================================================================
%=================================================================
%=================================================================
\newpage
\section{Anemia Use Case}\label{apd:anemia}
\subsection{Dataset Description}
\begin{figure}[htbp]
    \includegraphics[width=\linewidth]{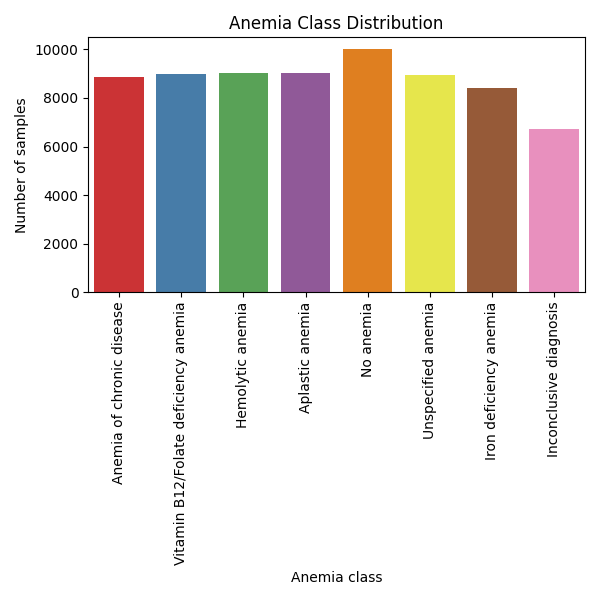}
    \caption{Number of patients per anemia class.}
    \label{fig:anem_class_distribution}
\end{figure}

\begin{figure}[htbp]
    \includegraphics[width=\linewidth]{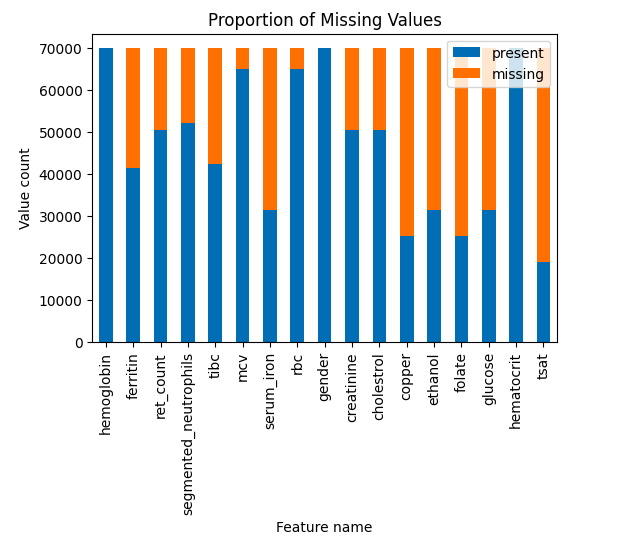}
    \caption{Number of present vs missing values for each feature in the \emph{anemia} dataset.}
    \label{fig:anem_missing_values}
\end{figure}

Figure \ref{fig:anem_class_distribution} shows the number of instances per diagnosis class in the synthetic dataset described in section \ref{subsec:anem_data_construction}. Figure \ref{fig:anem_missing_values} shows the ratio of missing versus observed values for each feature in the dataset. Additionally, Table \ref{tab:anem_sample} shows an example of a synthetic patient instance and Table \ref{tab:anem_statistics} shows a summarized statistical description of the features in the dataset. 

\begin{table*}[htbp]
    \centering    
    \begin{tabular}[]{@{}|l|l|@{}}
        \hline
         \bfseries Feature & \bfseries Value \\\hline
         \textbf{Hemoglobin} & 9.007012 \\
         \textbf{Ferritin} &  - \\
         \textbf{Reticulocyte count} & - \\
         \textbf{Segmented neutrophils} & 3.519565 \\
         \textbf{TIBC} & 440.499323 \\
         \textbf{MCV} & 103.442762 \\
         \textbf{Serum iron} & 59.017997 \\
         \textbf{RBC} & 2.612173 \\
         \textbf{Gender} & Male \\
         \textbf{Creatinine} & 0.650757 \\
         \textbf{Cholestrol} & 114.794964 \\
         \textbf{Copper} & 112.308159 \\
         \textbf{Ethanol} & 25.612786 \\
         \textbf{Folate} & 5.969710 \\
         \textbf{Glucose} & 116.026042 \\
         \textbf{Hematocrit} & 27.021037 \\
         \textbf{TSAT} & 13.397977 \\
         \textbf{label} & Vitamin B12/Folate deficiency anemia\\\hline
    \end{tabular}
    \caption{An instance in the \emph{anemia} dataset.}%
    \label{tab:anem_sample}
\end{table*}

\begin{table*}
    \fontsize{7pt}{7pt}\selectfont
    \centering
    \begin{tabular}[]{@{}|l|l|l|l|@{}}
    \hline
    \bfseries Feature & \bfseries All Classes & \bfseries No anemia & \bfseries \makecell{Vitamin B12/Folate \\ deficiency anemia} \\ \hline
    \textbf{Hemoglobin} &  10.239 (8.067, 12.102) & 14.570 (13.281, 15.848) & 9.510 (7.771, 11.274) \\
    \textbf{Ferritin} &  209.968 (69.870, 343.337) & 251.436 (128.145, 373.501) & 251.775 (126.676, 374.486) \\
    \textbf{Reticulocyte count} & 2.821 (1.270, 4.342) & 2.981 (1.520, 4.423) & 3.024 (1.526, 4.564)\\
     \textbf{Segmented neutrophils} & 2.930 (0.762, 4.898) & 3.532 (1.824, 5.266) & 3.516 (1.791, 5.22)\\
     \textbf{TIBC} & 334.276 (222.108, 457.942) & 310.821 (206.744, 415.958) &  306.678 (202.380, 410.634)  \\
     \textbf{MCV} & 89.998 (78.918, 101.093) & 90.029 (82.631, 97.598) & 102.504 (101.225, 103.794) \\
     \textbf{Serum iron} & 135.030 (77.793, 192.645) & 134.959 (76.979, 192.764) & 136.613 (80.087, 193.071) \\
     \textbf{RBC} & 3.348 (2.641, 3.936) & 4.899 (4.379, 5.356) & 2.784 (2.274, 3.298)\\
     \textbf{Gender} & & & \\
        \hspace{5mm}\textbf{Male} & 38268 (54.67\%) & 4048 (40.48\%) & 5108 (56.73\%) \\
        \hspace{5mm}\textbf{Female} & 31732 (45.33\%) & 5952 (59.52\%) & 3896 (43.27\%)\\
     \textbf{Creatinine} & 1.103 (0.651, 1.552) & 1.119 (0.670, 1.566) & 1.102 (0.655, 1.540)\\
     \textbf{Cholestrol} & 74.878 (37.388, 112.244) & 74.037 (36.257, 111.019) & 75.020 (37.099, 112.819) \\
     \textbf{Copper} & 80.095 (55.182, 105.245) & 79.510 (54.965, 104.816) & 80.001 (55.014, 105.150)  \\
     \textbf{Ethanol} & 39.887 (19.876, 59.749) & 40.256 (19.769, 60.865) & 39.445 (18.622, 59.590) \\
     \textbf{Folate} & 15.262 (7.832, 22.715) & 15.054 (7.843, 22.174) & 15.400 (8.099, 22.976) \\
     \textbf{Glucose} & 90.039 (65.128, 115.077) & 90.918 (66.287, 115.892) & 90.021 (64.290, 116.194)\\
     \textbf{Hematocrit} & 30.716 (24.201, 36.306) & 43.709 (39.843, 47.544) & 28.530 (23.313, 33.821) \\
     \textbf{TSAT} & 49.601 (23.103, 62.608) & 52.553 (24.626, 67.235) & 53.935 (25.712, 68.534) \\
  \hline
\end{tabular}

\bigskip

\begin{tabular}[]{@{}|l|l|l|l|@{}}
  \hline
    \bfseries Feature & \bfseries Unspecified anemia & \bfseries \makecell{Anemia of \\ chronic disease} & \bfseries Iron deficiency anemia \\\hline
         \textbf{Hemoglobin} & 9.534 (7.769, 11.301) & 9.514 (7.751, 11.254) & 9.539 (7.795, 11.276) \\
         \textbf{Ferritin} & 250.757 (123.567, 375.763) & 268.551 (152.371, 384.475) & 48.654 (22.742, 74.091) \\
         \textbf{Reticulocyte count} & 3.006 (1.546, 4.459) & 2.957 (1.458, 4.457) & 2.975 (1.441, 4.502) \\
         \textbf{Segmented neutrophils} &  0.000 (0.000, 0.000) & 3.580 (1.848, 5.284) & 3.582 (1.854, 5.330)\\
         \textbf{TIBC} & 311.332 (208.894, 414.291) & 301.558 (199.155, 402.335) & 452.223 (458.074, 499.101) \\
         \textbf{MCV} & 102.525 (101.274, 103.774) & 77.472 (76.206, 194.049) & 77.527 (76.299, 78.796)\\
         \textbf{Serum iron} & 134.970 (77.499, 192.216) & 135.313 (77.093, 194.049) &  135.625 (77.915, 193.272)\\
         \textbf{RBC} & 2.790 (2.274, 3.305) & 3.685 (3.008, 4.361)  & 3.693 (3.013, 4.371) \\
         \textbf{Gender}  & & & \\
            \hspace{5mm}\textbf{Male} & 5175 (57.73\%) & 5067 (57.20\%) & 4765 (56.74\%)\\
            \hspace{5mm}\textbf{Female} & 3789 (42.27\%) & 3792 (42.8\%) & 3633 (43.26\%) \\
         \textbf{Creatinine} & 1.096 (0.643, 1.554) & 1.098 (0.643, 1.544) & 1.103 (0.643, 1.558) \\
         \textbf{Cholestrol} & 75.734 (38.838, 113.730) & 74.994 (37.212, 112.230) & 74.700 (37.507, 112.034) \\
         \textbf{Copper} & 80.465 (56.729, 105.069) & 80.081 (54.461, 104.850) & 79.764 (54.906, 104.545)\\
         \textbf{Ethanol} & 39.500 (19.102, 22.784) & 39.784 (19.802, 59.703) & 39.501 (20.018, 58.701) \\
         \textbf{Folate} & 15.284 (7.919, 22.784) & 15.131 (7.769, 22.650) & 15.462 (7.948, 22.976)\\
         \textbf{Glucose} & 89.457(65.322, 113.808) & 90.131 (66.091, 115.144) & 90.153 (65.518, 114.407)\\
         \textbf{Hematocrit} & 28.602 (23.307, 33.902) & 28.541 (23.253, 33.761) & 28.617 (23.386, 33.828)\\
        \textbf{TSAT} & 52.689 (24.323, 67.385) & 54.538 (24.730, 71.416) & 32.675 (17.377, 42.591)\\\hline
\end{tabular}

\bigskip
% \subcaption*{The third classes}
% \begin{adjustbox}{width=\textwidth}
\begin{tabular}[]{@{}|l|l|l|l|@{}}
  \hline
    \bfseries Feature & \bfseries Hemolytic anemia & \bfseries Aplastic anemia & \bfseries \makecell{Inconclusive \\ diagnosis }\\\hline
    \textbf{Hemoglobin} & 9.510 (7.741, 11.262) & 9.518 (7.801, 11.250) & 9.486 (7.764, 11.224)\\
    \textbf{Ferritin} & 251.082 (123.988, 380.350) & 246.611 (119.831, 374.303) & 195.158 (67.362, 319.364)\\
    \textbf{Reticulocyte count} & 4.049 (3.079, 5.007) & 1.005 (0.514, 1.500) & 2.988 (1.437, 4.514)\\
     \textbf{Segmented neutrophils} & 3.553 (1.820, 5.271) & 3.517 (1.789, 5.189) & 3.478 (1.745, 5.214)\\
     \textbf{TIBC} & 310.197 (203.256, 418.237) & 310.744 (205.529, 414.594) & 338.282 (225.384, 461.215)\\
     \textbf{MCV} & 89.944 (84.920, 94.938) & 89.985 (85.003, 95.013) & 88.696 (78.100, 100.800)\\
     \textbf{Serum iron} & 133.068 (76.467, 191.047) & 135.668 (79.215, 192.992) & 133.704 (76.167, 191.807)\\
     \textbf{RBC} & 3.185 (2.580, 3.761)  & 3.186 (2.596, 3.747)  & 3.257 (2.628, 3.809)\\
     \textbf{Gender} & & & \\
        \hspace{5mm}\textbf{Male} & 5166 (57.24\%) & 5084 (56.31\%) & 3855 (57.36\%)\\
        \hspace{5mm}\textbf{Female} & 3859 (42.76\%) & 3945 (43.69\%) & 2866 (42.64\%)\\
     \textbf{Creatinine} & 1.098 (0.656, 1.538) & 1.100 (0.646, 1.551) & 1.108 (0.646, 1.563)\\
     \textbf{Cholestrol} & 74.968 (37.216, 112.418) & 74.990 (37.919, 112.314) & 74.602 (37.416, 111.479)\\
     \textbf{Copper} & 80.348 (55.201, 105.698) & 80.502 (55.372, 106.089) & 80.166 (55.087, 105.649)\\
     \textbf{Ethanol} & 40.220 (20.430, 60.389) & 40.525 (20.545, 60.523) & 39.744 (21.113, 58.165)\\
     \textbf{Folate} & 15.396 (7.854, 22.746) & 15.309 (7.731, 22.812) & 15.045 (7.521, 22.613)\\
     \textbf{Glucose} & 90.140 (65.137, 115.555) & 89.852 (64.575, 114.951) & 89.348 (63.492, 113.911)\\
     \textbf{Hematocrit} & 28.530 (23.223, 33.785) & 28.554 (23.402, 33.751) & 28.458 (23.292, 33.671)\\
     \textbf{TSAT} & 53.800 (24.902, 69.688) & 53.292 (26.181, 69.297) & 48.753 (21.476, 63.018)\\\hline
\end{tabular}   
\caption{Summary descriptive statistics of the \emph{anemia} dataset showing the mean and interquartile range (in parentheses) of the features. Gender, which is a binary variable, is described using the sample number and the percentage.}
 \label{tab:anem_statistics}
\end{table*}

\subsection{Sample Learned Pathways}
\begin{figure}[!ht] 
    \centering
    \subfloat[][]{\includegraphics[width=0.45\textwidth]{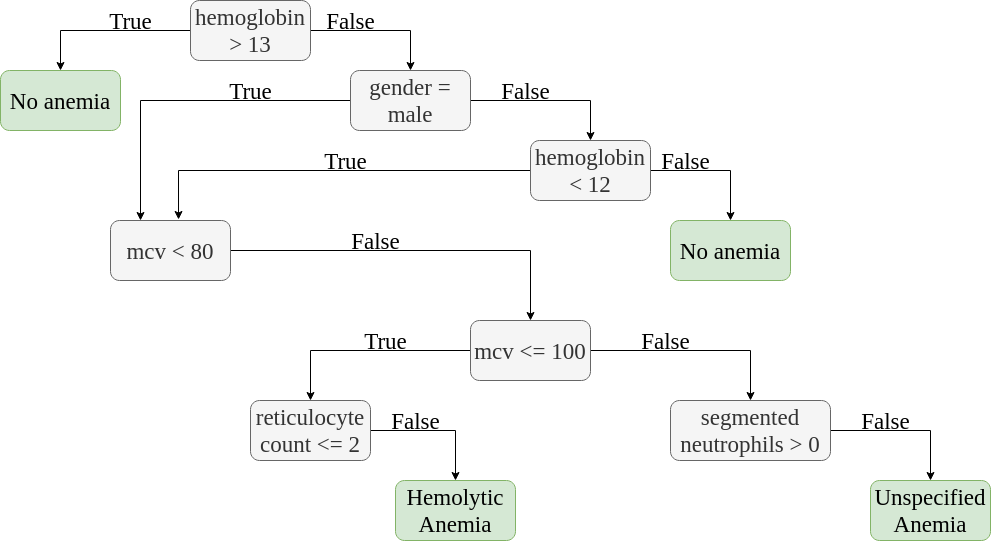}\label{fig:tree_path}} 
    \subfloat[][]{\includegraphics[width=0.45\textwidth]{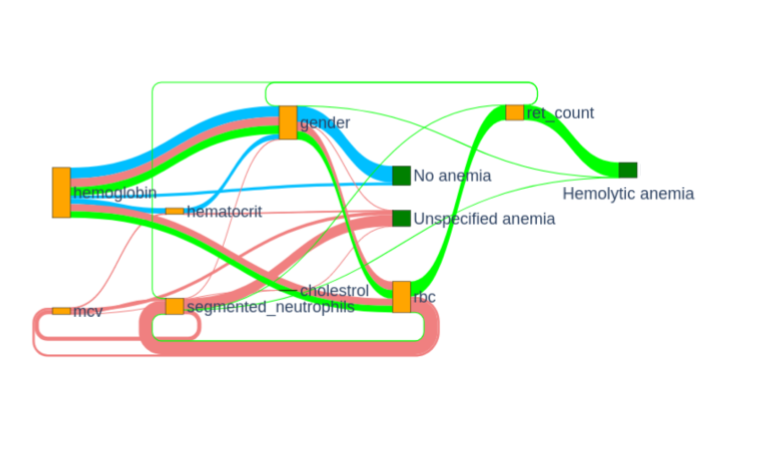}\label{fig:pathway_side}} 
    \caption{A side-by-side illustration of (\textit{a}) a branch of the decision tree in Figure \ref{fig:tree} and (\textit{b}) the corresponding pathways learned by the model for \emph{No anemia}, \emph{Hemolytic anemia} and \emph{Unspecified anemia}.}
    \label{fig:side_by_side}
\end{figure}

Figures \ref{fig:side_by_side} 
% and \ref{fig:pathway2} 
provides another example of pathways generated with our approach. 
% In particular, Figure \ref{fig:side_by_side} 
It displays a side-by-side illustration of a branch of the decision tree used to label the dataset and a Sankey diagram showing the pathways learned by the Dueling DQN-PER model for the diagnosis of solely \emph{No anemia}, \emph{Unspecified anemia} and \emph{Hemolytic anemia}, which are colored \emph{blue}, \emph{coral} and \emph{light green}, respectively. The feature actions are represented by the orange nodes, while the diagnosis actions are the dark green nodes. In addition, the size of each flow corresponds to its support, \textit{i.e.}, the number of patients that have it in their pathway.

% \begin{figure}[!ht] 
%     \centering
%     \subfloat[][]{\includegraphics[width=0.45\textwidth]{images/tree_4_pathway1.drawio.png}\label{fig:tree_path}} 
%     \subfloat[][]{\includegraphics[width=0.45\textwidth]{images/anem_pathway1.png}\label{fig:pathway_side}} 
%     \caption{A side-by-side illustration of (\textit{a}) a branch of the decision tree in Figure \ref{fig:tree} and (\textit{b}) the corresponding pathways learned by the model for \emph{No anemia}, \emph{Hemolytic anemia} and \emph{Unspecified anemia}.}
%     \label{fig:side_by_side}
% \end{figure}

\subsection{Computing Time}
\begin{table*}[!p]
\centering
    \begin{tabular}{|p{3.5cm}|p{4.4cm}|p{4.4cm}|}
    \hline
    \bfseries Model & \bfseries Training time & \bfseries Testing time\\
    \hline
         Decision Tree & 299 ms ± 9.72 ms  & 18.6 µs ± 453 ns \\ 
         Random Forest & 5.85 s ± 8.56 ms & 3.27 ms ± 156 µs \\
         SVM & 25.8 s ± 437 ms & 166 µs ± 8.33 µs \\ 
         FFNN & 2min 36s ± 35 s & 645 µs ± 28.6 µs \\ 
         Dueling DQN-PER & 1h 42 min 14s ± 19 min 19s & 722 µs ± 35.1 µs \\ 
         Dueling DDQN-PER & 2h 12min 45s ± 7min 36s & 759 µs ± 56.1 µs \\
    \hline
    \end{tabular}
    \caption{The computing time to train the model and to generate a diagnosis (pathway) for a single \emph{anemia} test instance.}
    \label{tab:anem_resources}
\end{table*}

In Table \ref{tab:anem_resources}, the time taken to train each model (training time), as well as, the time taken to generate a diagnosis pathway or a diagnosis (testing time) for the DQN models and the SOTA models respectively for a single instance in the test dataset are shown.

\newpage
\section{Lupus Use Case}\label{apd:lupus}

\subsection{Determining Penalty Weights}
Table \ref{tab:feature_scores} shows the penalty weights of the different features that were used to determine the penalty incurred by the agent on selecting that action as used in the reward function in \ref{eqn:lupus_feat_actions}. The scores were calculated based on the level of invasiveness of the action, the turnaround time for its observation, and the financial cost of performing that action. These values were derived from a review of existing literature and online sources. 
A breakdown of how the scores of each criterion were generated is shown in Table \ref{tab:feature_score_criterion}.

\begin{table}[!ht]
\centering
\resizebox{\textwidth}{!}{\begin{tabular}{|l|l|l|l|l|}
\hline
\textbf{Feature} & \textbf{Invasiveness} & \textbf{Turnaround time} & \textbf{Financial cost} & \textbf{Weighted score ($c$)} \\ 
\hline
ANA & 4 & 3 & 4 & 11.5\\
\hline
Fever & 5 & 5 & 5 & 15 \\
\hline
Leukopenia & 4 & 4 & 4 & 12 \\
\hline
Thrombocytopenia & 4 & 4 & 4 & 12 \\
\hline
Autoimmune hemolysis & 4 & 4 & 4 & 12 \\
\hline
Delirium & 5 & 5 & 5 & 15 \\
\hline
Psychosis & 5 & 5 & 5 & 15 \\
\hline
Seizure & 5 & 5 & 5 & 15 \\
\hline
Non-scarring alopecia & 5 & 5 & 5 & 15 \\
\hline
Oral ulcers & 5 & 5 & 5 & 15 \\
\hline
Cutaneous lupus  & 2 & 2 & 3 & 6.5 \\
\hline
Pleural effusion & 5 & 3 & 3 & 13 \\
\hline
Pericardial effusion & 5 & 3 & 3 & 13 \\
\hline
Acute pericarditis  & 5 & 3 & 3 & 13 \\
\hline
Joint involvement & 5 & 3 & 3 & 13 \\
\hline
Proteinuria & 5 & 3 & 4 & 13.5 \\
\hline
Renal biopsy-proven lupus nephritis & 0 & 2 & 0 & 1 \\
\hline
Anti-cardiolipin antibodies & 4 & 1 & 3 & 10 \\
\hline
Anti-$\beta$2GP1 antibodies  & 4 & 0 & 3 & 9.5\\
\hline
Lupus anti-coagulant & 4 & 0 & 3 & 9.5\\
\hline
Low C3 & 4 & 1 & 4 & 10.5 \\
\hline
Low C4 & 4 & 1 & 4 & 10.5 \\
\hline
Anti-dsDNA antibody & 4 & 3 & 4 & 11.5\\
\hline
Anti-Smith antibody & 4 & 3 & 4 & 11.5\\
\hline
\end{tabular}}
\caption{The penalty weights of the different features. The final score for each feature is a weighted number calculated by $2\times\text{invasiveness} + 0.5\times\text{turnaround time} + 0.5\times\text{financial cost}$. A breakdown of how the penalty weights for each criterion were generated is shown in Table \ref{tab:feature_score_criterion}.} % in Appendix \ref{apd:feature_scores}.}
\label{tab:feature_scores}
\end{table} 

\begin{table}[!ht]
\centering
    \begin{tabular}{|l|l|l|l|}
    \hline
    \textbf{Rating} & \textbf{Invasiveness} & \textbf{Turnaround time} & \textbf{Financial cost}  \\ \hline
    5 & none & none & none  \\ \hline
    4 & very low & within 1 day & < USD 50 \\ \hline
    3 & low & 1 - 3 days & USD 50 - 300 \\ \hline
    2 & medium & 3 - 7 days & USD 301 - 700 \\ \hline
    1 & high & 7 - 14 days & USD 701 - 1000 \\ \hline
    0 & very high & > 2 weeks & > USD 1000 \\ \hline
    \end{tabular}
    \caption{The values used to determine the penalty weights.}
    \label{tab:feature_score_criterion}
\end{table}

\subsection{Dataset Description}
Table \ref{tab:lupus_literature_fractions} shows the prevalence of the different features in patients with SLE as well as the literature source from which they were determined. An instance in the dataset is shown in Table \ref{tab:lupus_data_instance}, while a summary statistical description of the dataset is shown in Table \ref{tab:lupus_summary_statistics}.
\begin{table}
    \centering
    \begin{tabular}{|p{4cm}|p{2cm}|p{3cm}|p{3cm}|}
    \hline
        \textbf{Feature name} & \textbf{Literature source} & \textbf{Percentage from literature} & \textbf{Percentage used in dataset creation for SLE positive patients}\\
        \hline
        ANA & \cite{aringer20192019} & 100\% & 100\% \\
        \hline
        Fever & \cite{timlin2018fevers} & 36 - 86\% & 18\%\\
        \hline
        Leukopenia & \cite{fayyaz2015haematological}  & 50 - 60\% & 25\%\\
        \hline
        Thrombocytopenia & \cite{galanopoulos2017lupus} & 20 - 40\% &  10\%\\
        \hline
        Autoimmune hemolysis & \cite{crawford2022refractory}  & 3\% & 1.5\% \\
        \hline
        Delirium & \cite{nayak2012psychosis}  & 49.33\% & 24.5\%\\
        \hline
        Psychosis & \cite{nayak2012psychosis} & 12\% &  6\%\\
        \hline
        Seizure & \cite{nayak2012psychosis} & 10.67\% &  5.5\%\\
        \hline
        Non-scarring alopecia & \cite{desai2021recent} & 85\% & 42.5\% \\
        \hline
        Oral ulcers & \cite{kudsi2021prevalence} & 8 - 45\% & 4\% \\
        \hline
        Subacute cutaneous lupus  & \cite{gronhagen2014cutaneous} & 10\% & 5\% \\
        \hline
        Acute cutaneous lupus & \cite{gronhagen2014cutaneous} & 3\% & 1.5\% \\
        \hline
        Discoid lupus  & \cite{provost1994relationship} & 15 - 20\% & 7.5\% \\
        \hline
        Pleural effusion & \cite{yao2020clinical} & 17 - 60\% & 8.5\% \\
        \hline
        Pericardial effusion & \cite{almousa2022unusual} & 50\% & 25\% \\
        \hline
        Acute pericarditis  & \cite{narang2022acute} & 1\% & 0.5\% \\
        \hline
        Joint involvement & \cite{ceccarelli2017joint} & 69 - 95\% & 34.5\% \\
        \hline
        Proteinuria & \cite{nasim2022low}  & 60\% & 30\%\\
        \hline
        Renal biopsy-proven lupus nephritis & \cite{hong2018systematic}  & 16.9\% & 8.5\%\\
        \hline
        Anti-cardiolipin antibodies & \cite{unlu2016clinical} & 17 - 40\% & 8.5\% \\
        \hline
        Anti-$\beta$2GP1 antibodies  & \cite{dema2016autoantibodies} & 10 - 35\% & 5\% \\
        \hline
        Lupus anticoagulant & \cite{unlu2016clinical} & 11 - 30\% & 5.5\% \\
        \hline
        Low C3 & \cite{ramos2004hypocomplementemia}  & 45\% & 22.5\% \\
        \hline
        Low C4 & \cite{ramos2004hypocomplementemia} & 44\% & 22\% \\
        \hline
        Anti-dsDNA antibody & \cite{fabrizio2015systemic}  & 70 - 98\% & 35\% \\
        \hline
        Anti-Smith antibody & \cite{arroyo2015clinical}  & 9 - 49\% & 4.5\% \\
        \hline
    \end{tabular}
    \caption{The prevalence of the various features in SLE-positive patients in the \emph{lupus} dataset. The dataset utilized half the minimum percentage from the literature source to prevent a situation where nearly every positive ANA patient is also diagnosed with lupus. } 
    \label{tab:lupus_literature_fractions}
\end{table}

\begin{table}[]
    \centering
    \begin{tabular}[]{@{}|l|c|@{}}
    \hline
        \textbf{Feature} & \textbf{Value} \\
        \hline
         \textbf{ANA} & 1 \\
         \textbf{Fever} & 0  \\
         \textbf{Leukopenia} & 0 \\
         \textbf{Thrombocytopenia} & 1 \\
         \textbf{Autoimmune hemolysis} & 0\\
         \textbf{Delirium} & 1 \\
         \textbf{Psychosis} & 0 \\
         \textbf{Seizure} & 0 \\
         \textbf{Non-scarring alopecia} & 0 \\
         \textbf{Oral ulcers} & 0 \\
         \hline
         \textbf{Cutaneous lupus} & 0 \\
         \hline
         \textbf{Pleural effusion} & 0 \\
         \hline
         \textbf{Pericardial effusion} & 0 \\
         \hline
         \textbf{Acute pericarditis} & 0 \\
         \hline
         \textbf{Joint involvement} & 0 \\
         \hline
         \textbf{Proteinuria} & 1 \\
         \hline
         \textbf{Renal biopsy-proven lupus nephritis} & 0 \\
         \hline
         \textbf{Anti-cardiolipin antibodies} & 0 \\
         \hline
         \textbf{Anti-$\beta$2GP1 antibodies} & 0 \\
         \hline
         \textbf{Lupus anticoagulant} & 0 \\
         \hline
         \textbf{Low C3} & 0 \\
         \hline
         \textbf{Low C4} & 0 \\
         \hline
         \textbf{Anti-dsDNA antibody} & 0 \\
         \hline
         \textbf{Anti-Smith antibody} & 0 \\
         \hline
         \textbf{Label} & Lupus \\
        \hline
    \end{tabular}
    \caption{An instance in the \emph{Lupus} dataset.}
    \label{tab:lupus_data_instance}
\end{table}

\begin{table}
\centering
\resizebox{\textwidth}{.34\paperheight}{\begin{tabular}{|l|l|l|l|}
\hline
\textbf{Feature }& \textbf{All Classes} & \textbf{No lupus} & \textbf{Lupus} \\ 
\hline
ANA & \makecell[l]{0 - 20000 (28.6\%) \\ 1 - 70000(71.4\%)} & \makecell[l]{0 - 20000 (57.1\%) \\ 1 - 15056(42.9\%)} & \makecell[l]{0 - 0 (0\%) \\ 1 - 34944(100\%)} \\
\hline
Fever &  \makecell[l]{0 - 60274 (86.1\%) \\ 1 - 9726(13.9\%)} & \makecell[l]{0 - 32363 (92.3\%) \\ 1 - 2693 (7.7\%)} & \makecell[l]{0 - 27911 (79.9\%) \\ 1 - 7033(20.1\%)} \\
\hline
Leukopenia &  \makecell[l]{0 - 56776 (81.1\%) \\ 1 - 13224 (18.9\%)} & \makecell[l]{0 - 31929 (91.1\%) \\ 1 - 3127 (8.9\%)} & \makecell[l]{0 - 24847 (71.1\%) \\ 1 - 10097 (28.9\%)} \\
\hline
Thrombocytopenia & \makecell[l]{0 - 64388 (92.0\%) \\ 1 - 5612 (8\%)} & \makecell[l]{0 - 33712 (96.2\%) \\ 1 - 1344 (3.8\%)} & \makecell[l]{0 - 30676 (87.8\%) \\ 1 - 4268 (12.2\%)} \\
\hline
Autoimmune hemolysis &  \makecell[l]{0 - 68600 (98\%) \\ 1 - 1400 (2\%)} & \makecell[l]{0 - 34304 (97.9\%) \\ 1 - 752 (2.1\%)} & \makecell[l]{0 - 34296 (98.1\%) \\ 1 - 648 (1.9\%)} \\
\hline
Delirium &  \makecell[l]{0 - 56898 (81.3\%) \\ 1 - 13102 (18.7\%)} & \makecell[l]{0 - 31521 (89.9\%) \\ 1 - 3535 (10.1\%)} & \makecell[l]{0 - 25377 (72.6\%) \\ 1 - (27.4\%)} \\
\hline
Psychosis &  \makecell[l]{0 - 66335 (94.8\%) \\ 1 - 3665 (5.2\%)} & \makecell[l]{0 - 33840 (96.5\%) \\ 1 - 1216 (3.5\%)} & \makecell[l]{0 - 32495 (93.0\%) \\ 1 - 2449 (7.0\%)} \\
\hline
Seizure &  \makecell[l]{0 - 66762 (95.4\%) \\ 1 - 3238 (4.6\%)} & \makecell[l]{0 - 34277 (97.8\%) \\ 1 - 779 (2.2\%)} & \makecell[l]{0 - 32485 (93.0\%) \\ 1 - (7.0\%)} \\
\hline
Non-scarring alopecia &  \makecell[l]{0 - 48040 (68.6\%) \\ 1 - 21960 (31.4\%)} & \makecell[l]{0 - 29220 (83.4\%) \\ 1 - 5836 (16.6\%)} & \makecell[l]{0 - 18820 (53.9\%) \\ 1 - 16124 (46.1\%)} \\
\hline
Oral ulcers &  \makecell[l]{0 - 67247 (96.1\%) \\ 1 - 2753 (3.9\%)} & \makecell[l]{0 - 33792 (96.4\%) \\ 1 - 1264 (3.6\%)} & \makecell[l]{0 - 33455 (95.7\%) \\ 1 - 1489 (4.3\%)} \\
\hline
Cutaneous lupus &  \makecell[l]{0 - 61370 (87.7\%) \\ 1 - 3143 (4.5\%) \\ 2 - 1111 (1.6\%)\\ 3 - 4376 (6.3\%)} & \makecell[l]{0 - 32408 (92.4\%) \\ 1 - 1003 (2.9\%) \\ 2 - 447 (1.3\%)\\ 3 - 1198 (3.4\%)} & \makecell[l]{0 - 28962 (82.9\%) \\ 1 - 2140 (6.1\%) \\ 2 - 664 (1.9\%)\\ 3 - 3178 (9.1\%)} \\
\hline
Pleural effusion &  \makecell[l]{0 - 65212 (93.2\%) \\ 1 - 4788 (6.8\%)} & \makecell[l]{0 - 34043 (97.1\%) \\ 1 - 1013 (2.9\%)} & \makecell[l]{0 - 31169 (89.2\%) \\ 1 - 3775 (10.8\%)} \\
\hline
Pericardial effusion &  \makecell[l]{0 - 57098 (81.6\%) \\ 1 - 12902 (18.4\%)} & \makecell[l]{0 - 33075 (94.3\%) \\ 1 - 1981 (5.7\%)} & \makecell[l]{0 - 24023 (68.7\%) \\ 1 - 10921 (31.3\%)} \\
\hline
Acute pericarditis  &  \makecell[l]{0 - 69328 (99.0\%) \\ 1 - 672 (1.0\%)} & \makecell[l]{0 - 34633 (98.8\%) \\ 1 - 423 (1.2\%)} & \makecell[l]{0 - 34695 (99.3\%) \\ 1 - 249 (0.7\%)} \\
\hline
Joint involvement &  \makecell[l]{0 - 52195 (74.6\%) \\ 1 - 17805 (25.4\%)} & \makecell[l]{0 - 33248 (94.8\%) \\ 1 - 1808 (5.2\%)} & \makecell[l]{0 - 18947 (54.2\%) \\ 1 - 15997 (45.8\%)} \\
\hline
Proteinuria &  \makecell[l]{0 - 54516 (77.9\%) \\ 1 - 15484 (22.1\%)} & \makecell[l]{0 - 32138 (91.7\%) \\ 1 - 2918 (8.3\%)} & \makecell[l]{0 - 22378 (64.0\%) \\ 1 - 12566 (36.0\%)} \\
\hline
\makecell[l]{Renal biopsy-proven\\ lupus nephritis} &   \makecell[l]{0 - 64148 (91.6\%) \\ 1 - 1306 (1.9\%) \\2 - 954 (1.4\%) \\ 3 - 1089 (1.6\%) \\ 4 - 1829 (2.6\%) \\ 5 - 674 (1.0\%)} & \makecell[l]{0 - 33352 (95.1\%) \\ 1 - 1126 (3.2\%) \\2 - 211 (0.6\%) \\ 3 - 95 (0.3\%) \\ 4 - 83 (0.2\%) \\ 5 - 189 (0.5\%)} & \makecell[l]{0 - 30796 (88.1\%) \\ 1 - 180 (0.5\%) \\2 - 743 (2.1\%) \\ 3 - 994 (2.8\%) \\ 4 - 1746 (5.0\%) \\ 5 - 485 (1.4\%)} \\
\hline
Anti-cardiolipin antibodies &  \makecell[l]{0 - 64890 (92.7\%) \\ 1 - 5110 (7.3\%)} & \makecell[l]{0 - 33395 (95.3\%) \\ 1 - 1661 (4.7\%)} & \makecell[l]{0 - 31495 (90.1\%) \\ 1 - 3449(9.9\%)} \\
\hline
Anti-$\beta$2GP1 antibodies  & \makecell[l]{0 - 66581 (95.1\%) \\ 1 - 3419 (4.9\%)} & \makecell[l]{0 - 33703 (96.1\%) \\ 1 - 1353 (3.9\%)} & \makecell[l]{0 - 32878 (94.1\%) \\ 1 - 2066 (5.9\%)} \\
\hline
Lupus anticoagulant &  \makecell[l]{0 - 66424 (94.9\%) \\ 1 - 3576 (5.1\%)} & \makecell[l]{0 - 33640 (96.0\%) \\ 1 - 1416 (4.0\%)} & \makecell[l]{0 - 32784 (93.8\%) \\ 1 - 2160 (6.2\%)} \\
\hline
Low C3 &  \makecell[l]{0 - 57962 (82.8\%) \\ 1 - 12038 (17.2\%)} & \makecell[l]{0 - 32112 (91.6\%) \\ 1 - 2944 (8.4\%)} & \makecell[l]{0 - 25850 (74.0\%) \\ 1 - 9094 (26.0\%)} \\
\hline
Low C4 &  \makecell[l]{0 - 58221 (83.2\%) \\ 1 - 11779 (16.8\%)} & \makecell[l]{0 - 32179 (91.8\%) \\ 1 - 2877 (8.2\%)} & \makecell[l]{0 - 26042 (74.5\%) \\ 1 - 8902 (25.5\%)} \\
\hline
Anti-dsDNA antibody &  \makecell[l]{0 - 52168 (74.5\%) \\ 1 - 17832 (25.5\%)} & \makecell[l]{0 - 33107 (94.4\%) \\ 1 - 1949 (5.6\%)} & \makecell[l]{0 - 19061 (54.5\%) \\ 1 - 15883 (45.5\%)} \\
\hline
Anti-Smith antibody & \makecell[l]{0 - 67399 (96.3\%) \\ 1 - 2601 (3.7\%)} & \makecell[l]{0 - 34439 (98.2\%) \\ 1 - 617 (1.8\%)} & \makecell[l]{0 - 32960 (94.3\%) \\ 1 - 1984 (5.7\%)} \\
\hline
\end{tabular}}
\caption{Summary statistics of the \emph{Lupus} dataset showing the sample number and percentage of each value for each feature.}
\label{tab:lupus_summary_statistics}
\end{table} 

\subsection{Sample Learned Pathways}

\begin{figure}[htbp]
    \centering
    \includegraphics[width=\linewidth]{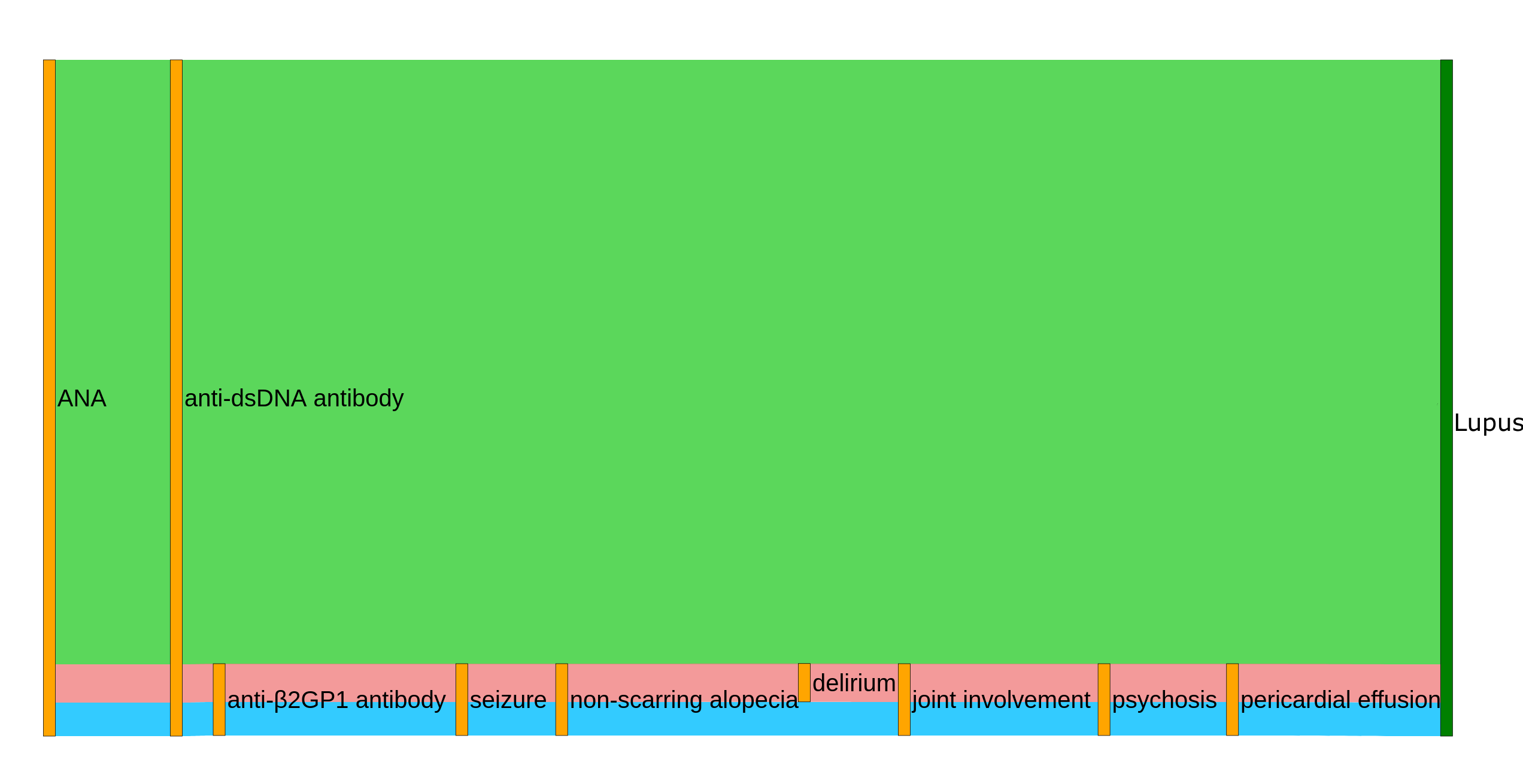}
    \caption{The three commonest diagnostic decision pathways for the \emph{Lupus} class generated by one of the models with the best wPAHM score. The model used to generate the pathways was the one with a wPAHM score closest to the mean wPAHM of the ten runs.}
    \label{fig:lupus_wpahm_lupus}
\end{figure}

Figure \ref{fig:lupus_wpahm_lupus} shows the three commonest pathways for the \emph{Lupus} class generated by one of our experiments: the dueling DDQN model chosen based on the wPAHM score. 

\begin{figure*}[!ht] 
    \centering
    \includegraphics[width=\linewidth]{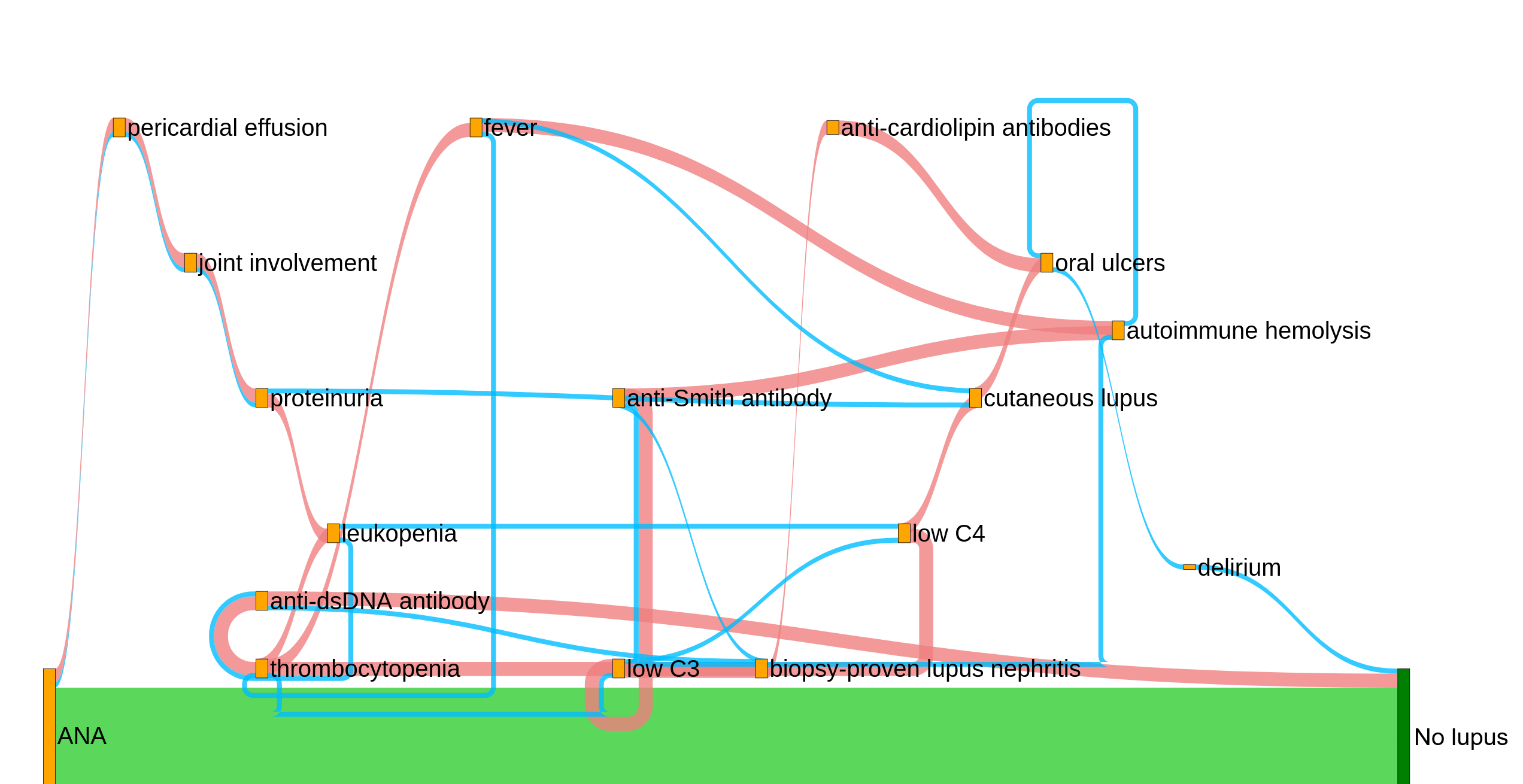}
    \caption{The three commonest diagnostic decision pathways for the \emph{No lupus} class generated by one of the models with the best accuracy score. The model used to generate the pathways was the one with an accuracy score closest to the mean accuracy of the ten runs.}
    \label{fig:lupus_acc_no_lupus}
\end{figure*}

\begin{figure*}[!ht] 
   \centering
    \includegraphics[width=\linewidth]{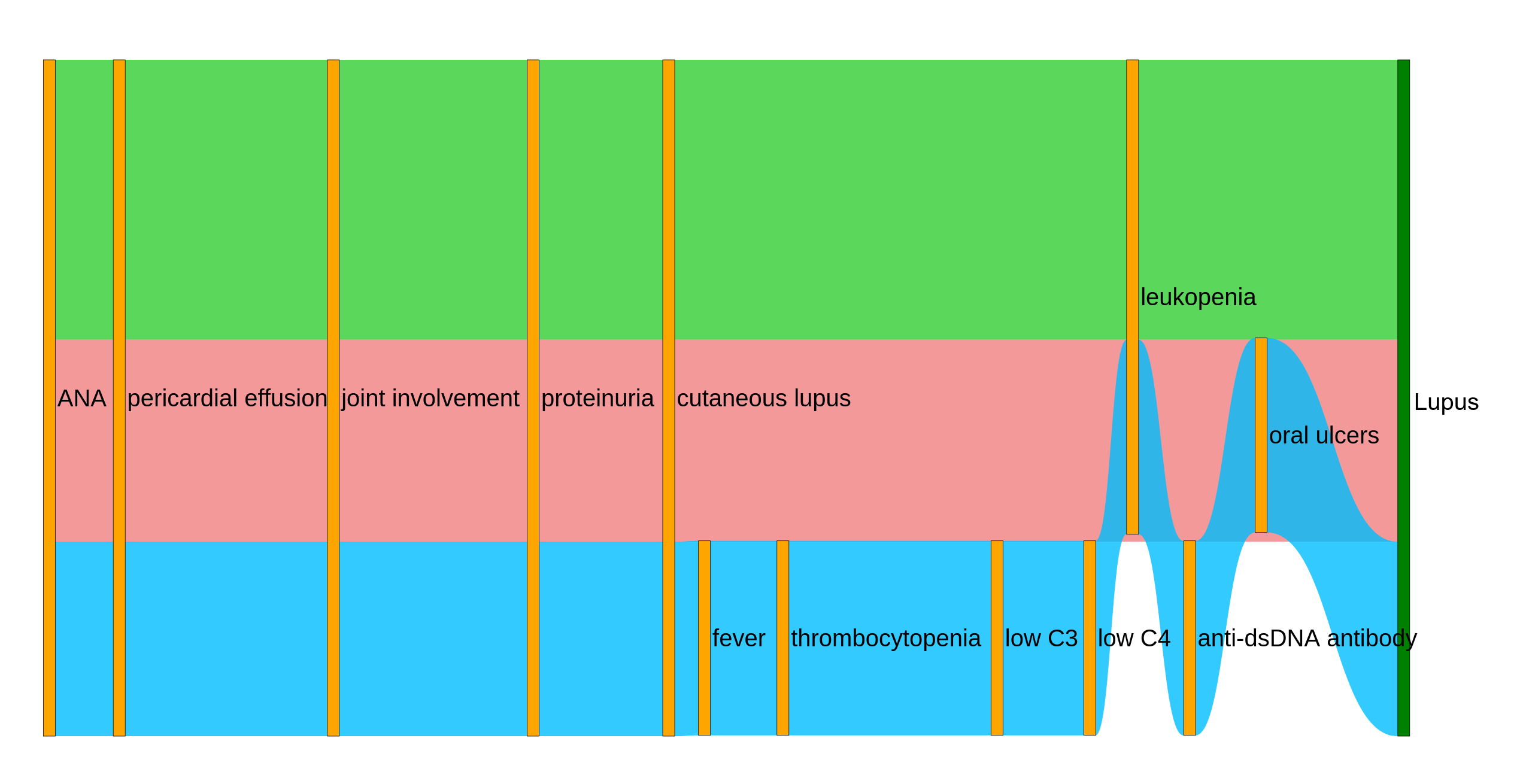}
    \caption{The three commonest diagnostic decision pathways for the \emph{Lupus} class generated by one of the models with the best accuracy score. The model used to generate the pathways was the one with an accuracy score closest to the mean accuracy of the ten runs.}
    \label{fig:lupus_acc_lupus}
\end{figure*}

Figures \ref{fig:lupus_acc_no_lupus} and \ref{fig:lupus_acc_lupus} show the three commonest pathways for the \emph{No lupus} and \emph{Lupus} classes respectively, generated by one of our experiments: the dueling DDQN model chosen based on the accuracy score. 

The models used were the ones with a score closest to the mean metric of the ten runs with $\lambda$ set to 9.

% Regarding the lupus use case, Figures \ref{fig:lupus_pathways} and \ref{fig:no_lupus_pathways} show the three commonest pathways for each class generated by one of our experiments: the dueling DDQN model chosen based on the best accuracy and wPAHM, with $\lambda$ set to 9.
\subsection{Computing Time}
\begin{table*}[!p]
\centering
    \begin{tabular}{|p{3.5cm}|p{5cm}|p{3.5cm}|}
    \hline
    \bfseries Model & \bfseries Training time & \bfseries Testing time\\
    \hline
         Decision Tree & 25.3 ms ± 991 µs  & 29.3 µs ± 771 ns \\ 
         Random Forest & 772 ms ± 46.8 & 3.16 ms ± 98.5 µs \\
         SVM & 3.09 s ± 35.1 ms & 114 µs ± 1.66 µs \\ 
         FFNN & 3min 21s ± 633 ms & 597 µs ± 22.9 µs \\ 
         Dueling DQN-PER & 13h 53min 35s ± 12min 10s & 1.79 ms ± 62.7 µs\\ 
         Dueling DDQN-PER & 15h 16min 38s ± 1h 49min 51s & 1.2 ms ± 12.6 µs\\
    \hline
    \end{tabular}
    \caption{The computing time to train the model and to generate a diagnosis (pathway) for a single \emph{lupus} test instance.}
    \label{tab:lupus_resources}
\end{table*}

In Table \ref{tab:lupus_resources}, the time taken to train each model (training time), as well as, the time taken to generate a diagnosis pathway or a diagnosis (testing time) for the DQN models and the SOTA models respectively for a single instance in the test dataset are shown.

%% else use the following coding to input the bibitems directly in the
%% TeX file.

% \begin{thebibliography}{00}

% %% \bibitem{label}
% %% Text of bibliographic item

% \bibitem{}

% \end{thebibliography}

\end{document}